\documentclass{article} 
\usepackage{collas2023_conference,times}
\usepackage{easyReview}

\usepackage{amsmath,amsfonts,bm}

\def\eqref#1{equation~\ref{#1}}

\def\1{\bm{1}}

\DeclareMathAlphabet{\mathsfit}{\encodingdefault}{\sfdefault}{m}{sl}
\SetMathAlphabet{\mathsfit}{bold}{\encodingdefault}{\sfdefault}{bx}{n}

\usepackage{hyperref}
\hypersetup{
    colorlinks=true,
    linkcolor=red,
    filecolor=magenta,
    urlcolor=blue,
    citecolor=purple,
    pdftitle={Overleaf Example},
    pdfpagemode=FullScreen,
    }

\definecolor{mydarkblue}{rgb}{0,0.08,0.45}

\usepackage{url}

\usepackage[utf8]{inputenc} 
\usepackage[T1]{fontenc}    
\usepackage{url}            
\usepackage{booktabs}       
\usepackage{amsfonts}       
\usepackage{nicefrac}       
\usepackage{microtype}      

\usepackage{sidecap}
\usepackage{float}
\usepackage{graphicx}

\usepackage{lineno}
\usepackage{multicol}
\usepackage{amsmath}
\usepackage{amssymb}
\usepackage{amsthm}
\usepackage{mathtools}
\usepackage[mathscr]{eucal}%
\usepackage{bm}
\usepackage{bbm}
\usepackage{relsize}
\usepackage{graphics}
\usepackage{wrapfig}
\usepackage{multirow}
\usepackage{enumitem}
\usepackage{tablefootnote}
\usepackage{array}
\usepackage{color, colortbl}
\definecolor{Gray}{gray}{0.9}
\usepackage{algorithm}
\usepackage{algorithmic}
\usepackage{pgf}
\usepackage{xinttools}
\usepackage{tikz}
\usetikzlibrary{arrows.meta}
\usepackage{textcomp}
\usetikzlibrary{calc,shapes,arrows,positioning,automata,trees}
\usepackage{placeins} 
\usepackage{tabularx}

\usepackage{graphicx}
\usepackage{subcaption} 
\usepackage{listings}

\newcommand{\norm}[1]{\left\lVert#1\right\rVert}

\theoremstyle{definition}

\newtheorem{theorem}{Theorem}

\usepackage{bigdelim}
\usepackage{colortbl} 
\definecolor{mColor1}{rgb}{0.95,0.95,0.95}

\usepackage{microtype}
\usepackage{graphicx}
\usepackage{booktabs} 

\usepackage{soul}

\definecolor{brown}{RGB}{150, 75, 0}
\definecolor{oldcolor}{RGB}{0,100,0}
\definecolor{newcolor1}{RGB}{255, 0, 0}
\definecolor{newcolor2}{RGB}{203,96,21}
\definecolor{newcolor3}{RGB}{25, 25, 112}
\definecolor{newcolor4}{RGB}{221, 160, 221}
\definecolor{newcolor5}{RGB}{48, 213, 200}

\usepackage[capitalise]{cleveref}

\Crefname{table}{Tab.}{Tabs.}
\Crefname{figure}{Fig.}{Figs.}
\Crefname{section}{Sec.}{Secs.}

\creflabelformat{equation}{#2\textup{(\textcolor{black}{#1})}#3}
\creflabelformat{table}{#2\textcolor{black}{#1}#3}
\creflabelformat{figure}{#2\textcolor{black}{#1}#3}
\creflabelformat{section}{#2\textcolor{black}{#1}#3}

\title{EMO: Episodic Memory Optimization for 
Few-Shot Meta-Learning}

\author{
Yingjun Du\textsuperscript{1},
Jiayi Shen\textsuperscript{1},
Xiantong Zhen\textsuperscript{1,2}\thanks{Currently with United Imaging Healthcare, Co., Ltd., China.}~,
Cees G. M. Snoek\textsuperscript{1} \\
\textsuperscript{1}AIM Lab, University of Amsterdam \textsuperscript{2}Inception Institute of Artificial Intelligence}

\collasfinalcopy 

\begin{document}

\maketitle

\begin{abstract}
Few-shot meta-learning presents a challenge for gradient descent optimization due to the limited number of training samples per task. 
To address this issue, we propose an episodic memory optimization for meta-learning, we call \emph{EMO}, which is inspired by  the human ability to recall past learning experiences from the brain's memory. EMO retains the gradient history of past experienced tasks in external memory, enabling few-shot learning in a memory-augmented way. 
By learning to retain and recall the learning process of past training tasks, EMO nudges parameter updates in the right direction, even when the gradients provided by a limited number of examples are uninformative.
We prove theoretically that our algorithm converges for smooth, strongly convex objectives. EMO is generic, flexible, and model-agnostic, making it a simple plug-and-play optimizer that can be seamlessly embedded into existing optimization-based few-shot meta-learning approaches. Empirical results show that EMO scales well with most few-shot classification benchmarks and improves the performance of optimization-based meta-learning methods, resulting in accelerated convergence.

\end{abstract}

\section{Introduction}

The vast majority of current few-shot learning methods fall within the general paradigm of meta-learning~\citep{schmidhuber1987evolutionary, bengio1990learning, thrun_metalearning}.  It searches for the best few-shot learning strategy as the learning experiences increase~\citep{finn17, ravi17, andrychowicz2016learning}. Optimization-based meta-learning~\citep{finn17, ravi17, li2017meta, raghu2019rapid} is one of the most popular approaches, owing to its ``model-agnostic'' nature to incorporate different model architectures and its principled formulation that allows the application to various problems. 
Optimization-based meta-learning comprises inner-loop and outer-loop updates that operate on a batch of tasks per iteration. In the inner-loop, these methods learn task-specific network parameters $\theta$ by performing traditional gradient descent on a task-specific loss $\mathcal{L}(\theta; \mathcal{S})$ with the support set $\mathcal{S}$, where
\begin{equation} \small
    \theta^{'} = \theta - \alpha  \nabla_{\theta} \mathcal{L}(\theta; \mathcal{S}),
\end{equation}
and $\alpha$ is the learning rate which determines the step size per inner iteration.
Gradient estimation with a small support set is inherently noisy, which causes the model to diverge or converge to a non-optimal minimum per task. 
Due to the small number of samples, traditional optimizers, e.g.,~\citep{allen2016improved, Adam, sutskever2013importance, ruder2016overview, li2017learning}, tend to get trapped in local minima. In this paper, we propose a new inner-loop optimizer for few-shot meta-learning.

Our work is inspired by the human cognitive function of episodic memory in the brain, which enables us to quickly adapt to new tasks with limited training samples by recalling past learning experiences from episodic memory~\citep{tulving1972episodic, tulving1983elements, tulving2002episodic}. Episodic memory has been shown to be effective in various machine learning tasks, such as reinforcement learning and continual learning. In reinforcement learning  \citep{zhu2020episodic, gershman2017reinforcement}, recent works use episodic memory to store past experiences and improve generalization ability quickly. In continual learning \citep{lopez2017gradient, chaudhry2019tiny}, episodic memory alleviates catastrophic forgetting while allowing beneficial knowledge transfer to previous tasks. Building on this inspiration, we introduce episodic memory into meta-learning for few-shot learning. Our approach learns to collect long-term episodic optimization knowledge for improved few-shot learning performance. By incorporating the cognitive function of episodic memory into meta-learning, we aim to improve machine learning models' generalization and adaptation ability to new tasks with limited training samples.

This paper proposes a new inner-loop optimization method for few-shot meta-learning, we call \emph{Episodic Memory Optimization} (EMO).  EMO exploits an external memory to accrue and store the gradient history gained from past training tasks, enabling the model to update to optimal parameters more accurately and quickly when faced with new tasks. 
Specifically, the episodic memory stores the gradients of the parameters per network layer for previous tasks, which are aggregated with the gradient of the current task in a linear or learnable way.  By doing so, episodic memory could help us achieve more optimal model parameters, despite having only limited training samples available for the new task. 
 To avoid overloading the memory storage space, EMO incorporates a memory controller that implements three different replacement strategies to replace the task in the memory.
Furthermore, we also prove that EMO with ﬁxed-size memory converges under strong convexity assumptions, regardless of which gradients are selected or how they are aggregated to form the update step.
EMO is a general gradient descent optimization that is model-agnostic and serves as a plug-and-play module that can seamlessly be embedded into existing optimization-based {few-shot} meta-learning approaches. 
We conduct our ablations and experiments on the few-shot learning benchmarks and verify that the optimization-based meta-learning methods with EMO easily outperform the original methods in terms of both performance and convergence.

\section{Method}

\subsection{Preliminaries}

\paragraph{Few-shot classification} The goal of few-shot classification is to construct a model using a limited number of labelled examples. In the conventional few-shot classification scenario, following~\citep{vinyals16}, we define the $N$-way $K$-shot classification problem, which has $N$ classes, and each class has $K$ labelled support examples. In this scenario each task is a classification problem from a predefined task distribution $p(\mathcal{T})$. We denote the labeled support set by $\mathcal{S} {=} \{(x_{i}, y_{i})\}_{i=1}^{N\times K}$;  each $(x_{i}, y_{i})$ is a pair of an input and a label, where $y_i\in\{1,2,\cdots,N\}$.   Each task is also associated with a query set $\mathcal{Q} {=} \{(x_{j}, y_{j})\}_{j=1}^{N \times M}$  
to evaluate the quality of the trained model.
The query set $\mathcal{Q}$ for each task is also composed of examples of the same $N$ classes. 
Usually, optimizing and learning parameters for each task with a few labelled training samples is difficult. Meta-learning offers a way of learning to improve performance by leveraging knowledge from multiple tasks.

\paragraph{Optimization-based meta-learning} 
In meta-learning, a sequence of tasks  $\{\mathcal{T}_1, \mathcal{T}_2, \cdots, \mathcal{T}_{N_{\mathcal{T}}}\}$ are sampled from a predefined task distribution $p(\mathcal{T})$, where each one is a few-shot learning task.
The core idea of meta-learning is to find a well-generalized meta-learner on the training tasks during the meta-training phase. For each task \begin{small}$\mathcal{T}_i$\end{small}, the meta-learner \begin{small}$\Phi$\end{small} is applied on the base learner \begin{small}$f_{\theta_{i}}$\end{small}, and the parameter \begin{small}$\theta_{i}$\end{small} and meta-learner \begin{small}$\Phi$\end{small} are learned alternatively. 
During the meta-testing phase, the learned meta-learner is applied to tackle the new tasks composed of examples from unseen classes. Given a new few-shot learning task \begin{small}$\mathcal{T}_t$\end{small}, the optimal meta-learner \begin{small}$\Phi^{*}$\end{small} is used to improve the effectiveness of \begin{small}$\mathcal{T}_t$\end{small} by solving \begin{small}$\min_{\theta_t}\mathcal{L}(\Phi^{*}(f_{\theta_{t}}), \mathcal{Q})$\end{small}. In this way,  meta-learning effectively adapts to new tasks, even when the training data for the new task is insufficient. 
Optimization-based meta-learning~\citep{finn17, li2017learning, raghu2019rapid} strives to learn an optimization that is shared across tasks while being adaptable to new tasks. The most representative optimization-based meta-learning is model-agnostic meta-learning (MAML) by~\cite {finn17}. MAML is formulated as a bilevel optimization problem with inner-loop and outer-loop optimization, where the inner-loop computes the task-specific parameters $\theta^{'}$ (starting from $\theta$) via a few gradient updates:
   \begin{equation} \label{eq:maml_inner}
	\theta_{t+1} = 
	\theta - 
	\alpha \nabla_\theta 
	\frac{1}{N \times K} 
	\sum\limits_{(x, y) \in \mathcal{S}}
	\mathcal{L}_{\mathcal{T}_i}(f_{\theta_t}(x), y) \ .
\end{equation} 
For the outer loop, the original model parameters are then updated  after the inner-loop update, i.e.,
\begin{equation} \label{eq:maml_outer}
	\theta \leftarrow 
	\theta - 
	\beta \nabla_\theta 
	\frac{1}{N_{\mathcal{T}}} 
	\sum\limits_{\{\mathcal{T}_1, \cdots, \mathcal{T}_t\}}
	\frac{1}{M \times K}
	\sum\limits_{(x, y) \in \mathcal{Q}}
	\mathcal{L}_{\mathcal{T}_i}(f_{\theta_{t+1}}(x), y) \ .
\end{equation}
where $\alpha$ and $\beta$ are the inner-loop and outer-loop learning rates, respectively. Training results in a model initialization $\theta$ that can be adapted to any new task with just a few gradient steps.

\subsection{Model}
This paper focuses on inner-loop optimization for optimization-based meta-learning methods.  We propose a new inner-loop optimization method called \emph{Episodic Memory Optimization} (EMO).  
Our proposed EMO model is composed of four parts: 
an \textbf{encoder} that generates a representation for the incoming query data and the available support data;
an external \textbf{memory store} which contains previously seen task representations and the gradients of each layer with writing managed by a \textbf{memory controller}; 
and an \textbf{episodic gradient memory stochastic gradient descent} that ingests the gradients from the new task and data from the memory store to generate the new gradients over the current task. 

\paragraph{Encoder} The encoder first takes in support data  $\mathcal{S} {=} \{(x_{i}, y_{i})\}_{i=1}^{N\times K}$ and then converts these data to the representation $\{e_{i}\}_{i=1}^{N\times K}$ of lower dimension. In this paper, the input is an image, and we choose a convolutional network architecture for the encoder function $f_{\theta}$.  

\paragraph{Memory store} Our  external memory module $\mathcal{M} {=} \{M_{t}\}_{t=1}^{T}$ contains the stored learning process of experienced tasks, where $T$ is the memory capacity. Each of the slots corresponds to each experienced task. In our work, the memory stores a key-value pair in each row of the memory array as~\citep{graves_nmt}. The keys are the task representations  $K_t$ of each task, the value is the gradient representation as value $V_t^l$, $M_t = [\mathbf{K}_t, \mathbf{V}_t]$, where $\mathbf{V}_t = \{V_t^1, V_t^2, \cdots, V_t^l\}$,  $t$ indicates the task $t$ and $l$ indicates the  $l$-th convolutional layer. The memory module is queried by finding the k-nearest neighbors between the test task representation and the task $K_t$ in a given slot. The distance metric used to calculate proximity between the points is an available choice, and here we always use Euclidean distance.  

For the task representation $K_t$, to allow the ﬂexibility of variable input sizes of task representations, we use the generic Transformer architecture~\citep{vaswani2017attention}:
\begin{equation}
    K_t = \texttt{Transformer}([\texttt{cls}_t, e_1, e_2, \cdots, e_n])[0],
\label{eqn:key}
\end{equation}
where $\texttt{cls}_t$ is the task representation token embedding, and $e_i = \textbf{Encoder}(x_i)$ is the encoded $i$-th support data pair $\mathcal{S} {=} \{(x_{i}, y_{i})\}_{i=1}^{N\times K}$. After the transformer, we take the position output \texttt{cls}   as the task embedding $K_t$. 

For the memory value $V_t^l$, we first compute the gradients of task $t$ at layer $l$ as:
\begin{equation}
   \mathbf{g}_{t}^l = \sum^{N \times K}_{i=1} \frac{\partial \mathcal{L}(\hat{{y}}_i^t, {y}_i^t)}{\partial \theta^l},
\label{eqn:gradients}
\end{equation}
where $\theta^l$ uses the parameters at layer $l$, we denote with $\mathcal{L}(\cdot)$ a loss function (such as the cross entropy loss on the labels). To avoid confusion, we omit the superscript $l$ for the memory from now on.

\paragraph{Memory controller} 
To avoid overloading the memory storage space, we propose a memory controller that decides to replace the episodic memory slot at a certain moment. The input of the memory controller consists of the gradient $\mathbf{g}_t$ of the current task and the selected memory $\hat{M}_{c}$ that needs to be replaced. The controller is written as:
\begin{align}
    M_c = \texttt{Controller} (\mathbf{g}_t, \hat{M}_c).
\end{align}
Inspired by the page replacement algorithm in operating systems, we propose three implementations of the memory controller:
\textit{First In First Out Episodic Memory} (\texttt{FIFO-EM}),
\textit{Least Recently Used Episodic Memory} (\texttt{LRU-EM}) and 
\textit{Clock Episodic Memory} (\texttt{CLOCK-EM}).
(1) \texttt{FIFO-EM} keeps track of all the  memory in a queue, with the oldest memory in the front. When a memory needs to be replaced, the memory in the front of the queue is selected for removal.  
(2) \texttt{LRU-EM} is a content-based memory writer that writes episodic memories to the least recently used memory location. 
New task information is written into rarely-used locations, preserving recently encoded data or written to the last used location, which can function as an update of the memory with newer, possibly more relevant information.
(3) In the 
\texttt{CLOCK-EM}, the candidate memory for removal is considered in a round-robin fashion, and a memory that has been accessed between consecutive considerations will be spared, similar to the CLOCK page replacement algorithm in the operating system~\citep{janapsatya2010dueling}. 
When the memory is $\mathcal{M}$ is not complete, we directly store the $\mathbf{g}_t$ to be added into $\mathcal{M}$, while once the memory is complete, we use the Controller to achieve memory replacement. The best-suited memory controller is specific to the underlying meta-learning method and datasets. We compare each $\texttt{Controller}$ in the experiments.

\paragraph{Episodic gradient memory stochastic gradient descent}
Episodic gradient memory stochastic gradient descent  is the explicit integration of episodic memory gradients into SGD.  To be specific, the iteration comes in the form: 
\begin{equation}
\label{eq:emo_inner}
    \theta_{t+1} = \theta_t - \alpha \cdot \texttt{Aggr}(\mathbf{g}_t,\ \mathcal{V}_t),
\end{equation}
where $\mathbf{g}_t$ are the gradients of the support set from the current task, $\mathcal{V}_t$ is the collection of  episodic gradients selected from the memory based on  the similarity of memory key, and the current task representation and \texttt{Aggr} denotes an aggregation function which is used to combine the episodic gradients with the current-iteration gradient.  We consider three possible functions for \texttt{Aggr} including \texttt{Mean}, the average of $\mathcal{V}_t$ and all selected episodic gradients;  \texttt{Sum}, the addition of $\mathcal{V}_t$ to the average of all selected episodic gradients; and \texttt{Transformer}, the learnable combination of $\mathcal{V}_t$ to  all the selected episodic gradients. Mathematically, these three \texttt{Aggr} functions are defined as: 
  \begin{equation}
    \texttt{Mean}(\mathbf{g}_t,\ \mathcal{V}_t) = \frac{1}{M_{\mathcal{V}_t}+1} (\mathbf{g}_t + \sum_{V_t \in \mathcal{V}_t} V_t),
\end{equation}\break
  \begin{equation}
    \texttt{Sum}(\mathbf{g}_t,\ \mathcal{V}_t) = \mathbf{g}_t + \frac{1}{M_{\mathcal{V}_t}} \sum_{V_t \in \mathcal{V}_t} V_t,
\end{equation}\break
  \begin{equation}
    \texttt{Transformer}(\mathbf{g}_t,\ \mathcal{V}_t) = \texttt{Transformer}([\texttt{cls}_g, \mathbf{g}_t, V_t^1, V_t^2, \cdots,  V_t^{M_{\mathcal{V}_t}}]) [0].
\end{equation}
where $\texttt{cls}_g$ is the new gradients token embedding.
The best-suited aggregation function is specific to the meta-learning method into which the episodic gradients are integrated. We compare each aggregation function with different optimization-based meta-learning methods in the experiments.

\subsection{Meta-training and meta-test}
Following~\citep{ravi17, finn17}, we perform episodic training by exposing the model to a variety of
tasks from the training distribution $p(\mathcal{T})$. For a given training task $\mathcal{T}_i$, the model first computes its parameters by Eq.~(\ref{eq:emo_inner}) in the inner-loop,  then incurs a loss $\mathcal{L}_i$ of this task, and updates the model parameters by Eq.~(\ref{eq:maml_outer}); we sum these losses and back-propagate through the sum at the end of the task. We evaluate the model using a partition of the dataset that is class-wise disjoint from the training partition. The parameters of EMO for each component are trained in an end-to-end framework.
In the meta-test stage, the model first computes the gradients $\mathbf{g}_t$ and recalls the memory $\mathcal{V}_t$ based on the task representation $k$, which is computed by the support set $\mathcal{S}$. Then the model updates 
the task-specific parameters by Eq.~(\ref{eq:emo_inner}) in the inner-loop.
After the inner-loop, we evaluate the model on the query set.  
Note that during the meta-test phase, our model only uses the content of the acquired memory to update the network parameters and does not modify the content stored in the memory. {Detailed algorithms for meta-training and meta-test are shown in the appendix~\ref{alg:EMO_train} and ~\ref{alg:EMO_test}.}

\subsection{Analysis of Convergence}
The core of EMO is to explicitly integrate the current gradient with the episodic memory $\texttt{Aggr}(\mathbf{g}_t, \mathcal{V}_t)$. In practice, we observe that the proposed method has a higher convergence rate than previous optimizers. Here, we theoretically analyze the proposed EMO optimization's convergence rate of gradient descent.  

To do so, we reformulate the aggregation process as a linear multi-step system~\citep{polyak1964some, assran2020convergence}, leading to $\texttt{Aggr}(\mathbf{g}_t, \mathcal{V}_t) {=} \sum_{s=0}^{S-1} w_{t, s} g_{t-s}$. $S$ is the number of steps. At the $t$-th iteration, the multi-step system involves the gradients from the past $S$ time steps. $w_{t, s}$ is the aggregation scalar of the corresponding gradient $g_{t-s}$ in the linear multi-step system, which is bounded by the interval $[0, 1]$. 
The system involves all gradients in the episodic memory, $\mathcal{V}_t  \subseteq\{ g_{t-s} \}_{s=0}^{S}$. 
For the gradient that does not appear in the memory, $\exists~s \in \{1, 2, ..., S\},  g_{t-s} \notin \mathcal{V}_t$, the corresponding aggregation scalar $w_{t, s}$ is $0$. 
In general, we define a model-agnostic objective as $\min_{\theta} f(\theta)$.  $\theta_t$ and $\theta^*$ denote model parameters of the $t$-th iteration and the optimal. The difference between both parameters is $\Delta \theta_t = \theta_t - \theta^*$. We assume $f$ is continuously differentiable, $\mu$-strongly convex and $L$-Lipschitz ($0 < \mu \leq L$). These assumptions imply the Hessian matrix  $\nabla^2 f(\theta)$ exists and is bounded by the interval $[\mu, L]$. We consider the stochastic gradient $g_t$ as a random vector and $\mathbb{E}[g_t] = \nabla f(\theta_t)$. $\epsilon_t$ denotes the independent gradient noise at iteration $t$. The gradient noise has zero means, and its variance is bounded by a finite constant $\sigma^2$. Thus, the gradient  in each iteration  can be formulated as:
   \begin{equation}
        g_{t} = \Delta \theta_t \int_{0}^{1} \nabla^2 f(\theta^* + u\Delta \theta_t)d u + \epsilon_t, 
\label{eq:formulation of gradient}
\end{equation} 
where $\int_{0}^{1} \nabla^2 f(\theta^* + u\Delta \theta_t)d u$ is the average rate of the gradient changes from the $t$-th iteration to the optimal one with respect to the model parameters. Based on the assumptions of the objective, the average rate is also bounded between $\mu$ and $L$. We incorporate Eq.~(\ref{eq:formulation of gradient}) into Eq.~(\ref{eq:emo_inner}) with the linear multi-step system. In this case, the convergence of the system depends on the spectral properties of the system matrix~\citep{mcrae2021memory}. 
\begin{theorem}[Convergence rate of EMO]
\label{thm: convergence}
We define a system matrix\footnote{For clarity, we provide the definition of  $A_t$ in Eq.~(\ref{eq:matrix vertion}) of Appendix~\ref{sec:proof}.}
for each iteration as $A_t$, which contains aggregation scalars and average rates of gradient changes of the past $S$ gradients. $\lambda_t$ is the square root of the largest singular value of the corresponding system matrix, and thus the spectral norm of the system matrix is not larger than $\lambda_t$. $\lambda_{\max}$ is the upper bound for all $\lambda_{t}$ corresponding to all system matrices. Since $\alpha$ is chosen sufficiently small such that $\lambda_{\max} < 1$, we have that:
\begin{equation}
\label{eq:upper_bound_gradient}
        f(\theta_{t+1}) - f(\theta^*) 
        \le \frac{L}{2}(\lambda_{\max}^{2t} \norm{ \Delta \theta_{1} }^2 + \frac{\alpha^2 \sigma^2 S }{1-\lambda^2_{\max}}).
\end{equation}
\end{theorem}
From this theorem, the learning rate mainly depends on $\lambda_{\max}$. The lower $\lambda_{\max}$, the faster the convergence rate and the smaller the variance.
When the number of steps in the system is $1$, Eq.~(\ref{eq:upper_bound_gradient}) degenerates to the conventional stochastic gradient decent as used in the previous meta-learning methods~\citep{finn17, li2017meta, raghu2019rapid}. In practice, our model usually sets a large number as the number of steps. In this case, it is possible to set the learning rate and aggregation scalars to obtain a faster convergence rate than SGD. Proofs are presented in Appendix~\ref{sec:proof}.

\section{Related Work}

\paragraph{Episodic memory}
Episodic memory has shown its effectiveness in a variety of machine learning tasks.
Recent works \citep{zhu2020episodic, gershman2017reinforcement, botvinick2019reinforcement, hu2021generalizable, lampinen2021towards} use episodic memory to store past experiences to help the intelligence quickly adapt to new environments and improve its generalization ability.   In continual learning, episodic memory alleviates catastrophic forgetting \citep{lopez2017gradient, chaudhry2019tiny, derakhshani2021kernel} while allowing beneficial transfer of knowledge to previous tasks.    We draw inspiration from the cognitive function of episodic memory and introduce it into meta-learning to learn to collect long-term episodic (optimization) knowledge for few-shot learning. 

\paragraph{Meta-learning} 
Meta-learning designs models to learn new tasks or adapt to new environments quickly with only a few training examples. There are four common research lines of meta-learning: (1) metric-based meta-learning ~\citep{snell2017prototypical,vinyals16,yang2018learning, du2021hierarchical, triantafillou2019meta} generally learn a shared/adaptive embedding space in which query images can be accurately matched to support images for classiﬁcation;
(2) optimization-based meta learning~\citep{finn17,finn2018probabilistic,lee2018gradient,yoon2018bayesian,grant2018recasting, kalais2022stochastic, abbas2022sharp, flennerhag2021bootstrapped, zou2021unraveling, triantafillou2019meta} learns an optimization algorithm that is shared across tasks and can be adapted to new tasks, enabling learning to be conducted efﬁciently and effectively. { Note that Proto-MAML~\citep{triantafillou2019meta} combines the strengths of prototypical networks and MAML for few-shot learning.  Proto-MAML initializes the task-specific linear layer from the ProtoNet before optimizing those parameters using MAML. Our method focuses on optimization-based meta-learning as well, with the key difference being our optimizer can be used with any meta-learning approach. }
; (3) model-based meta-learning~\citep{mishra2018simple,gordon2019meta} explicitly learns a base-learner that incorporates knowledge acquired by the meta-learner and effectively solves individual tasks; (4) memory-based meta-learning~\citep{munkhdalai2017meta, ramalho2019adaptive, zhen2020learning, santoro2016meta, du2021hierarchical, zhen2020icml} deploys an external memory to rapidly assimilate new data of unseen tasks, which is used for quick adaptation or to make decisions. Our method combines optimization-based meta-learning with memory-based meta-learning. 
To the best of our knowledge, it is the first optimization-based meta-learning method with episodic memory, intending to perform few-shot classiﬁcation.

\paragraph{Memory-based few-shot learning} 
 Both \cite{andrychowicz2016learning} and \cite{ravi17} propose the update rule for neural network parameters by transforming gradients via an LSTM, which outperforms fixed SGD update rules. The Meta-network~\citep{munkhdalai2017meta} learns to transform the gradients to fast weights as memory, which are stored and retrieved via attention during testing.
 Conditionally shifted neurons~\citep{munkhdalai2018metalearning}  modify the activation values with task-specific shifts retrieved from an external memory module, which is populated rapidly based on limited task experience.
 \citet{santoro2016meta} leverages the Neural Turning Machine \citep{graves_nmt} for online few-shot learning by designing efficient read-and-write protocols.
 \citet{ramalho2019adaptive} introduced adaptive posterior learning, which approximates probability distributions by remembering the most surprising observations it has encountered in external memory.
 \citet{babu2021online} proposed a distributed memory architecture, which recasts
the problem of meta-learning as simply learning with memory-augmented models.  
These methods~\citep{andrychowicz2016learning, ravi17} leverage an LSTM to design a new update rule for the network parameters, which can be seen as implicit memory. 
Compared to previous methods that rely on additive feature augmentation, our approach utilizes episodic memory to augment the gradients during the network parameter updating process. This represents a novel and distinct memory approach to few-shot learning.

\section{Experiments}
\subsection{Experimental Setup}

In our experiments we consider two datasets: (i) \textit{Meta-Dataset-BTAF}~\citep{yao2019hierarchically}, which contains four ﬁne-grained image classiﬁcation datasets: (a) \textit{Bird}~\citep{wah2011caltech}, \textit{Texture}~\citep{cimpoi2014describing}, \textit{Aircraft}~\citep{maji2013fine}, and \textit{Fungi}~\citep{fungi}. (ii) \textit{mini}ImageNet~\citep{vinyals16} which consists of 100 randomly chosen classes from ILSVRC2012~\citep{russakovsky15}.
For the \textit{Meta-Dataset-BTAF}, each meta-training and meta-test task samples classes from one of the four datasets. This benchmark is more heterogeneous and closer to real-world image classiﬁcation.  Following the conventional meta-learning settings~\citep{vinyals16, finn17}, all datasets are divided into meta-training, meta-validation and meta-testing classes. The $N$-way $K$-shot settings are used to split the training and test sets for each task. We report the average few-shot classiﬁcation accuracy (\%, top-1) along with the 95\% conﬁdence intervals across all test images and tasks. {The error bar in the Figure~\ref{fig:steps} and Figure~\ref{fig:memory_size}  represent the 95\% confidence intervals across all test images and tasks.  Appendix~\ref{sec:app_imp} provides the detailed implementation and algorithm.  The results  for MAML, ANIL, and Meta-SGD on the \textit{Meta-Dataset-BTAF} are based on our re-implementations.}

\subsection{Results}

\paragraph{Benefit of episodic memory optimizer}
\begin{table}[t]
\caption{Benefit of episodic memory optimizer for few-shot fine-grained classification. All evaluated optimization-based meta-learning methods consistently achieve better performance with EMO than without. Meta-SGD with EMO achieves the best performance, especially for the 5-way 5-shot setting. }
\label{tab:ab_EMO}
\centering
\scalebox{0.99}{
\begin{tabular}{lcccccc}
\toprule
& \multicolumn{2}{c}{MAML}                                  & \multicolumn{2}{c}{ANIL}                                  & \multicolumn{2}{c}{Meta-SGD} 
\\
\cmidrule(lr){2-3} \cmidrule(lr){4-5} \cmidrule(lr){6-7} 
Dataset                     & \multicolumn{1}{c}{w/o EMO} & \multicolumn{1}{c}{w/  EMO} & \multicolumn{1}{c}{w/o EMO} & \multicolumn{1}{c}{w/  EMO} & \multicolumn{1}{c}{w/o EMO} & \multicolumn{1}{c}{w/  EMO} \\
\midrule
\rowcolor{Gray}
5-way 1-shot & & & & & &\\
Bird 
& 53.94 \scriptsize{$\pm$ 1.45}                       
& 56.32 \scriptsize{$\pm$ 1.33}                      
& 52.57 \scriptsize{$\pm$ 1.44}                       
& 54.78 \scriptsize{$\pm$ 1.43}                     
& 55.58 \scriptsize{$\pm$ 1.43}                      
& {58.95} \scriptsize{$\pm$ 1.41}             \\

Texture                                       & 31.66 \scriptsize{$\pm$ 1.31}                      & {34.75} \scriptsize{$\pm$ 1.41}                      & 31.45 \scriptsize{$\pm$ 1.32}                      &33.15 \scriptsize{$\pm$ 1.31}                     & 32.38 \scriptsize{$\pm$ 1.32}                   &{36.26} \scriptsize{$\pm$ 1.33}             \\

Aircraft                                      & 51.37 \scriptsize{$\pm$  1.38}                       & {53.99}  \scriptsize{$\pm$  1.33}                       &  50.45 \scriptsize{$\pm$ 1.34}                       & {52.79} \scriptsize{$\pm$ 1.33}                       &  {52.99} \scriptsize{$\pm$  1.36}                      & {55.29} \scriptsize{$\pm$  1.35}             \\

Fungi                                         & 42.12 \scriptsize{$\pm$  1.36}                       &   {43.15} \scriptsize{$\pm$  1.36}                    & 41.14 \scriptsize{$\pm$  1.34}                        & {43.75} \scriptsize{$\pm$  1.31}                      & 41.74 \scriptsize{$\pm$  1.34}                      & {45.24} \scriptsize{$\pm$  1.34} \\       
\midrule
\rowcolor{Gray}
5-way 5-shot & & & & & &\\
Bird                                          &  68.52 \scriptsize{$\pm$ 0.79}                     &  {70.91} \scriptsize{$\pm$ 0.71}                     & 67.17 \scriptsize{$\pm$ 0.74}                       &  69.25 \scriptsize{$\pm$ 0.73}                      & 67.87 \scriptsize{$\pm$  0.74}                      & {72.74} \scriptsize{$\pm$  1.40}            \\

Texture                                       & 44.56 \scriptsize{$\pm$ 0.68}                      & 47.21 \scriptsize{$\pm$  0.64}                       &  43.41 \scriptsize{$\pm$ 0.68}                     &  45.78 \scriptsize{$\pm$ 0.68}                     & 45.49 \scriptsize{$\pm$ 0.68}                   &{49.15} \scriptsize{$\pm$ 0.68}            \\

Aircraft                                      & 66.18 \scriptsize{$\pm$  0.71}                       &  68.13 \scriptsize{$\pm$  0.61}                       &  65.34 \scriptsize{$\pm$ 0.70}                       & 67.15 \scriptsize{$\pm$ 0.71}                      & 66.84 \scriptsize{$\pm$  0.70}                      &  {69.73} \scriptsize{$\pm$  0.70}             \\

Fungi                                         & 51.85 \scriptsize{$\pm$ 0.85}                        &    56.17 \scriptsize{$\pm$  0.75}                   & 52.11 \scriptsize{$\pm$ 0.83}                         & 54.35 \scriptsize{$\pm$ 0.83}                       & 52.51 \scriptsize{$\pm$  0.81}                     & {58.21} \scriptsize{$\pm$  0.79} \\   
\bottomrule
\end{tabular}}
\end{table}
To show the benefit of our proposed episodic memory optimizer, we compare MAML~\citep{finn17}, Meta-SGD~\citep{li2017learning}, and ANIL~\citep{raghu2019rapid} with their EMO variants. 
Each original meta-learning method uses SGD as the inner-loop optimizer, while each EMO variant uses EMO as the inner-loop optimizer. 
Table \ref{tab:ab_EMO} shows adding EMO improves performance independent of the meta-learning method or dataset.  On the challenging Texture dataset, which has the largest domain gap, Meta-SGD with EMO delivers $36.26\%$, surpassing Meta-SGD by $3.88\%$.  In addition, Meta-SGD with EMO achieves the best performance compared with other meta-learning methods.
This is because Meta-SGD with EMO stores not only the gradients of each layer, but also the gradients of the learning rate in the inner-loop, thus  accelerating training. 
ANIL only stores the gradients of the last layer, causing the number of parameters and the memory size to be much smaller than in MAML and Meta-SGD. Despite the reduced accuracy, ANIL with EMO is still beneficial for applications that require compute efficiency, as ANIL is about 4.8 times faster than MAML and Meta-SGD.
We attribute the improvements with EMO to our model's ability to leverage the episodic memory to adjust the model parameters, allowing  the model to update the test task model using the most similar training task-like update.

\paragraph{Comparison with the other optimizers}
\begin{figure} [t]
  \begin{minipage}{1.0\columnwidth}
    \subfloat[Bird]{
      \includegraphics[width=.25\columnwidth]{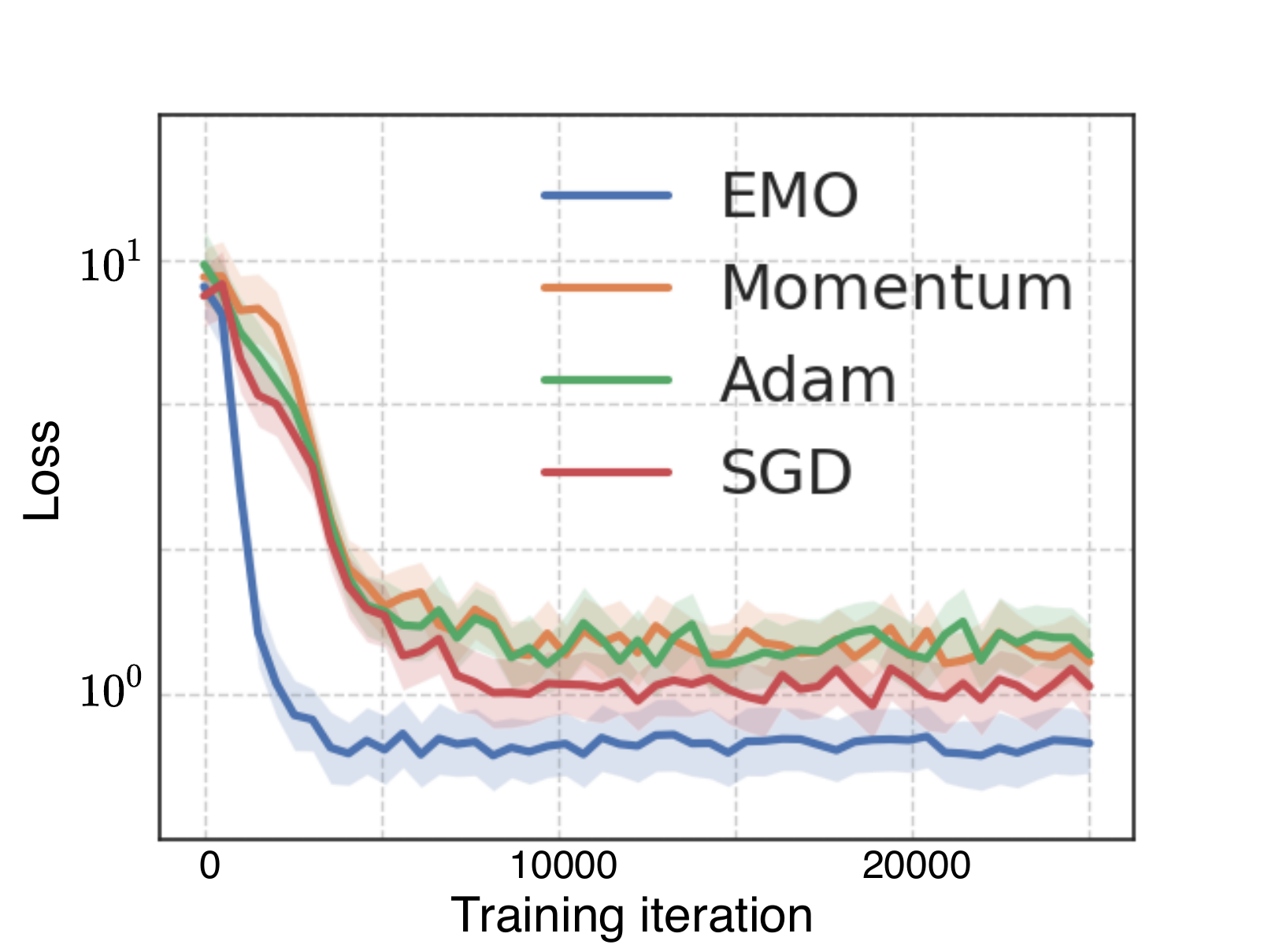}}
         \subfloat[Texture]{
      \includegraphics[width=.25\columnwidth]{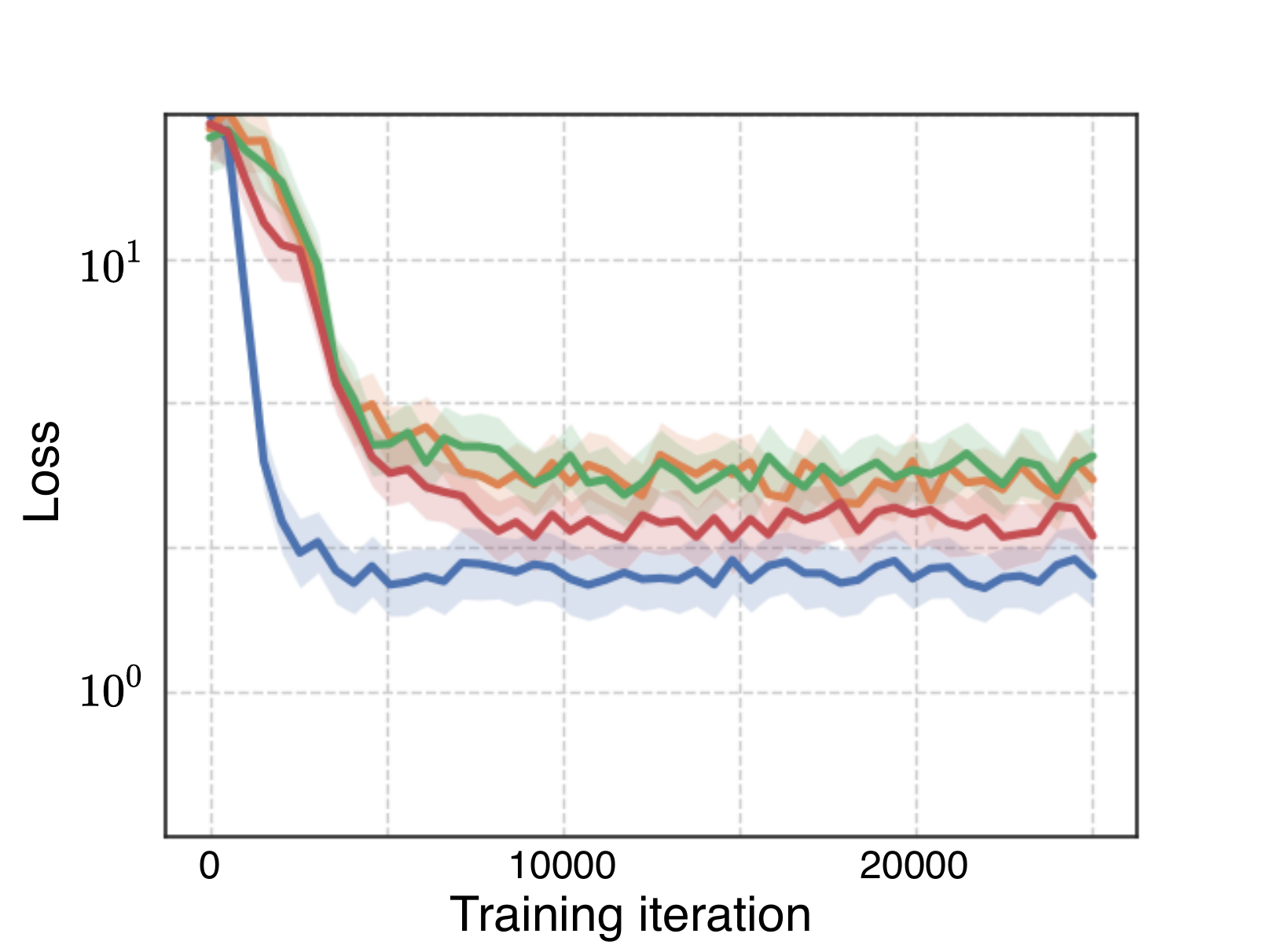}}
       \subfloat[Aircraft]{
      \includegraphics[width=.25\columnwidth]{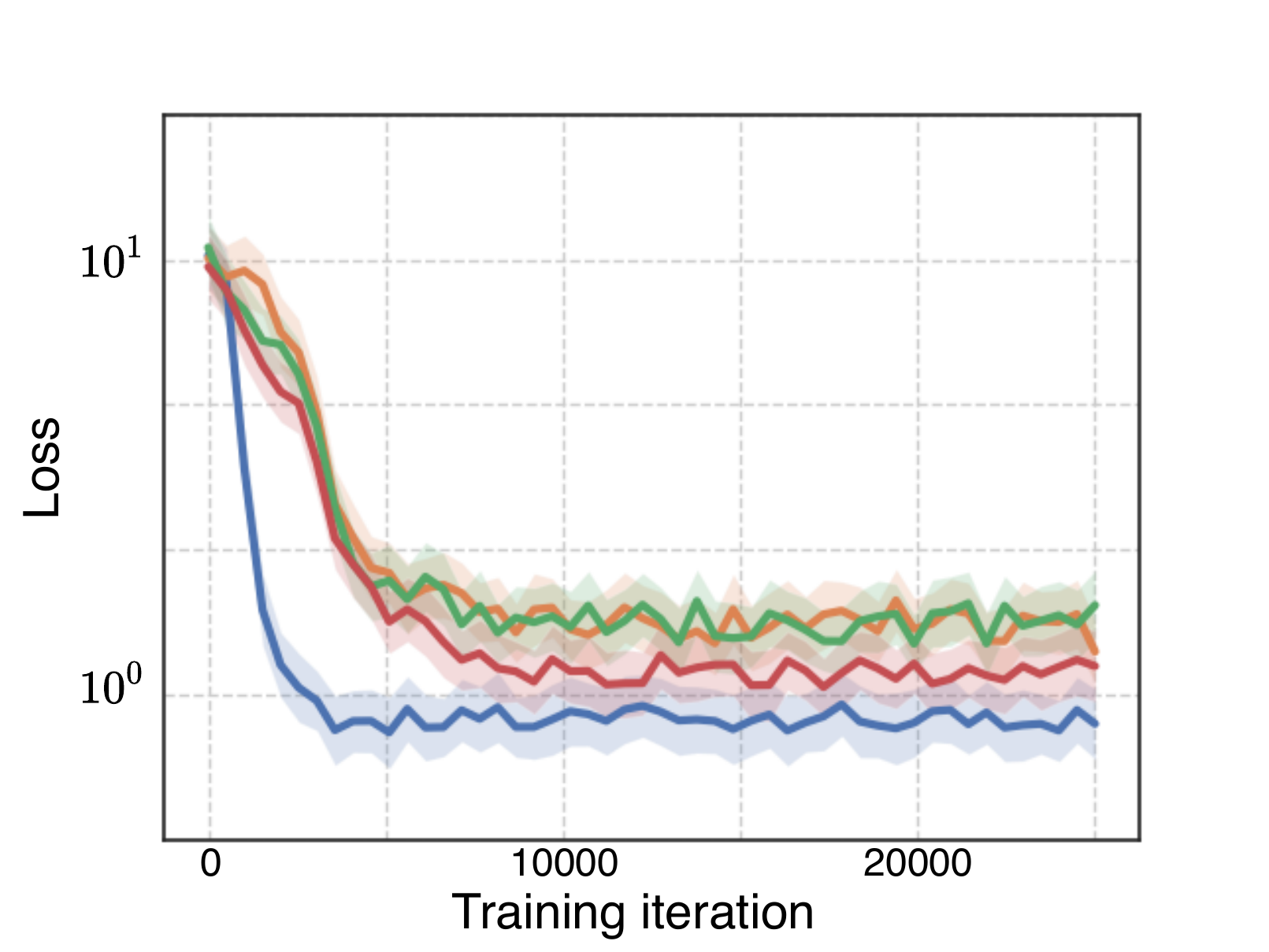}}
       \subfloat[Fungi]{
      \includegraphics[width=.25\columnwidth]{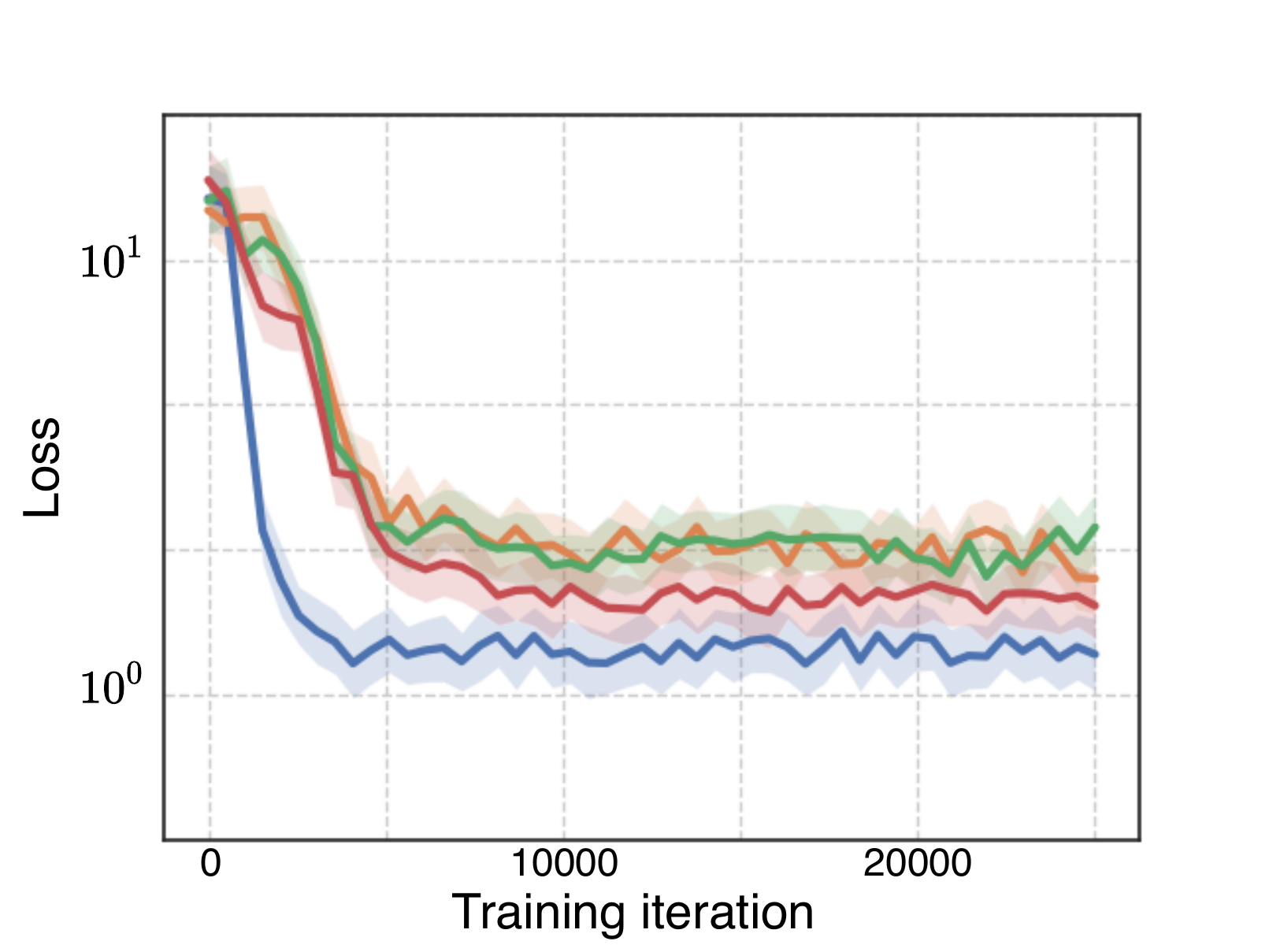}}
\end{minipage}
\caption{Comparisons for MAML with EMO and other optimizers.  EMO speeds up MAML training and outperforms the other optimizers for few-shot learning. }
	\label{fig:emo_loss}
\end{figure}
To show the benefit of our episodic memory optimizer, we compare EMO with other commonly used optimizers in the inner-loop stage of MAML. 
Learning curves for MAML using different optimizers are shown in Figure~\ref{fig:emo_loss}. 
EMO outperforms other optimizers by a considerable margin.  
Momentum and Adam have somewhat degraded performance compared to SGD, which means that these traditional optimizers cannot exploit past inaccurate gradients for few-shot learning. However, EMO can speed up training and improve performance since EMO acquires the ability to adaptively choose the most relevant task update rules for the test task.

\paragraph{Effect of inner-loop steps}

\begin{figure} [t]
\small
\centering
  \begin{minipage}{0.45\columnwidth}
      \includegraphics[width=0.8\columnwidth]{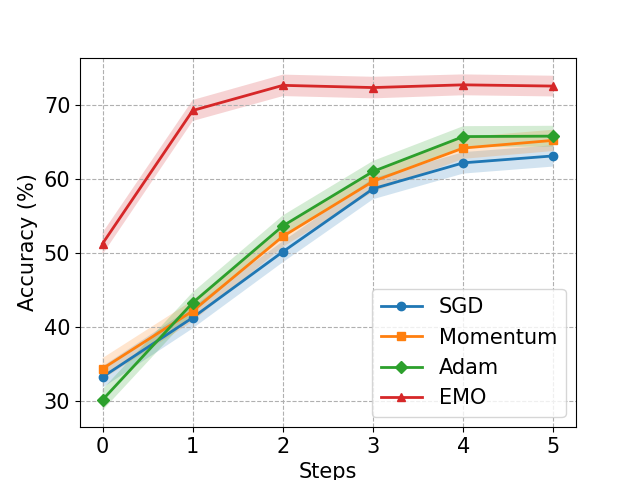}
      \centering
      \caption{Effect of inner-loop steps. }
      \label{fig:steps}
      \end{minipage}
  \begin{minipage}{0.45\columnwidth}
      \includegraphics[width=0.8\columnwidth]{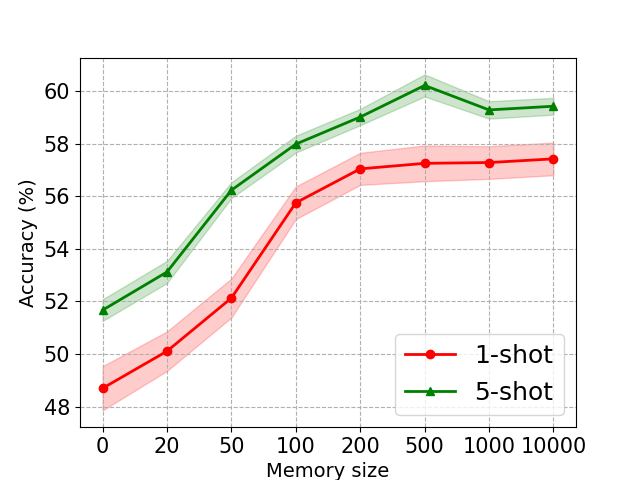}
      \centering
       \caption{Effect of task-memory size. }
       \label{fig:memory_size}
          \end{minipage}
\end{figure}
We provide further analysis of the effectiveness of our optimization in a fast adaptation by 
varying the number of update steps. Specifically, we compare the  performance of MAML with SGD, Momentum, Adam, and EMO in Figure~\ref{fig:steps}. We find that MAML with EMO achieves  about $51.34\%$ accuracy at step 0 (no support set is required to update the model), which is more than $19.86\%$ higher than MAML, since  EMO can utilize the past gradients in the memory to update the learning process of new tasks.  Also, MAML with EMO can reach convergence very quickly (step 2 vs. step 5) and perform much better than other optimizers.  The results by comparing them with different optimizers in the  Appendix~\ref{sec:other_optimizers}. We also report the adaptation speed of the MAML with different optimizers by the varying number of update steps in the Appendix~\ref{sec:speed}.
Although MAML with other optimizers already performs fast adaptation with 5 steps, MAML with EMO is even faster and better. This again demonstrates the benefit of EMO.

\paragraph{Comparison of our aggregation functions}
\begin{table}[t]
\caption{Effect of different aggregation functions on \textit{Meta-Dataset-BTAF} under the 5-way 1-shot setting.  The best-suited aggregation function for MAML is \texttt{Mean}, while  the best-suited aggregation function for  Meta-SGD is \texttt{Transformer}. 
}
\label{tab:ab_aggre}
\centering
\scalebox{0.9}{
\begin{tabular}{lcccccc}
\toprule
& \multicolumn{3}{c}{MAML with EMO}                     & \multicolumn{3}{c}{Meta-SGD with EMO} 
\\
\cmidrule(){2-4} \cmidrule(lr){5-7} 
Dataset                     & \multicolumn{1}{c}{ \texttt{Sum}} & \texttt{Mean}& {\texttt{Transformer}}    & {\texttt{Sum}} & {\texttt{Mean}}& {\texttt{Transformer}}  \\
\midrule
Bird 
&  54.35 \scriptsize{$\pm$ 1.34} 
&  56.32 \scriptsize{$\pm$ 1.33}                    
&  55.91 \scriptsize{$\pm$ 1.35}                   
&  57.15 \scriptsize{$\pm$ 1.31}                 
&  57.03 \scriptsize{$\pm$ 1.40}                   
&  {58.95} \scriptsize{$\pm$ 1.41}     
\\

Texture
& 33.13 \scriptsize{$\pm$ 1.45}  
& 34.75 \scriptsize{$\pm$ 1.41}                     
&  34.23 \scriptsize{$\pm$ 1.40}                   
&  34.93 \scriptsize{$\pm$ 1.42}                   
&   35.97 \scriptsize{$\pm$ 1.41}                   
&    {36.26} \scriptsize{$\pm$ 1.33}  
\\

Aircraft 
& 52.53  \scriptsize{$\pm$  1.30} 
& 53.99  \scriptsize{$\pm$  1.33} 
& 53.15  \scriptsize{$\pm$  1.30}                   
&     53.12 \scriptsize{$\pm$  1.27}                   
&     54.01 \scriptsize{$\pm$  1.25}                 
&    {55.29} \scriptsize{$\pm$  1.35}  
\\

Fungi  
&   44.07 \scriptsize{$\pm$  1.33}   
&   43.15 \scriptsize{$\pm$  1.36}                   
&  {45.27} \scriptsize{$\pm$  1.35}                  
&   43.49 \scriptsize{$\pm$  1.32}                   
&   44.13 \scriptsize{$\pm$  1.31}                    
&   45.24 \scriptsize{$\pm$  1.34} 
\\

\bottomrule
\end{tabular}}
\end{table}
We also ablate the effect of EMO's aggregation function to generate the new gradients. We report the performance of MAML and Meta-SGD with EMO using different \texttt{Aggr} in Table~\ref{tab:ab_aggre}, and the experiments for ANIL with EMO are proposed in Appendix~\ref{sec:ANIL_agg}. The results show that the best-suited aggregation function is specific to the optimization-based meta-learning method for integrating episodic gradients. The best-suited aggregation function for MAML with EMO is the \texttt{Mean}, while the best-suited aggregation function for Meta-SGD with EMO is the \texttt{Transformer}. To ensure consistency of implementation on each dataset and for each model, we choose the \texttt{Mean} aggregation function for MAML with EMO and the \texttt{Transformer} aggregation function for Meta-SGD with EMO in the remaining experiments.

\paragraph{Comparison of our memory controllers}
To assess the effect of the memory controller, we compare
our three memory controllers: \texttt{FIFO-EM}, \texttt{CLOCK-EM}, \texttt{LRU-EM} on the \textit{Meta-Dataset-BTAF} under the 5-way 1-shot setting.  The experimental results for MAML with EMO are
reported in Table~\ref{tab:controller}, and results for Meta-SGD and ANIL with EMO are in Appendix~\ref{sec:ANIL_controller}. \texttt{FIFO-EM} achieves the worst performance compared to the other memory controllers since  \texttt{FIFO-EM} may replace some crucial 
or commonly used memory, causing the test task to fail to find the precise memory to learn quickly. With \texttt{LRU-EM}, MAML with EMO 
leads to a small but consistent gain under all the datasets, as it replaces the 
memory that is not commonly used and these memories can usually be seen as
\begin{wraptable}{r}{7cm}
\vspace{-3mm}
\caption{Effect of the memory controller. \texttt{LRU-EM} achieves better performance than alternatives. } 
\label{tab:controller}
\centering
\scalebox{0.8}{\begin{tabular}{lccc}
\toprule
& \multicolumn{3}{c}{MAML with EMO} \\
\cmidrule(lr){2-4} 
Dataset & \texttt{FIFO-EM} & \texttt{CLOCK-EM} & \texttt{LRU-EM} \\
\midrule
Bird & 51.91 \scriptsize{$\pm$   1.35} &{54.01} \scriptsize{$\pm$   1.33} & {56.32} \scriptsize{$\pm$  1.33} 
\\
Texture & 30.11 \scriptsize{$\pm$   1.40} & {32.14} \scriptsize{$\pm$   1.41} & {34.75} \scriptsize{$\pm$   1.41}\\
Aircraft & 48.16 \scriptsize{$\pm$   1.40} & 50.91 \scriptsize{$\pm$   1.38} & {53.99} \scriptsize{$\pm$   1.33}\\
Fungi & 41.17 \scriptsize{$\pm$   1.35} & {43.97} \scriptsize{$\pm$   1.35} &  {43.15} \scriptsize{$\pm$   1.36}\\
\bottomrule
\end{tabular}}
\vspace{-3mm}
\end{wraptable}
outliers.  In Table \ref{tab:MSGD_controller}, 
with \texttt{CLOCK-EM}, Meta-SGD with EMO achieves better performance on the
all datasets. \texttt{CLOCK-EM} allows the network to access the memory in a systematic and efficient manner by controlling the sequence of read and write operations, which is more suitable for methods that require large memory, such as Meta-SGD with EMO, which also requires additional storage of the gradient of the inner-loop learning rate.
To ensure consistency of implementation on each dataset, we choose the \texttt{LRU-EM} function for MAML with EMO and ANIL with EMO, \texttt{CLOCK-EM} is used for Meta-SGD with EMO.

\paragraph{Effect of task-memory size}
Task-memory size cannot be increased indefinitely.  To study the effect 
We conduct this ablation of task-memory size on EMO on \textit{mini}ImageNet using MAML with EMO under the 5-way 1-shot and 5-shot settings. {Note that task memory size is the number of stored meta-training tasks. For each task, we store the model gradients and their task representation.}
From Figure~\ref{fig:memory_size}, we observe the performance increases along with the increase in task-memory size. This is expected since more significant memory provides more context information for building better memory. Naturally, the memory size has a greater impact in the 1-shot setting. In this case, the model updated from only one example might be insufficiently representative of the object class. Leveraging context information provided by the memory compensates for the limited number of samples. We adopt memory sizes 100 for 1-shot and 200 for 5-shot on each dataset.

\paragraph{Computational cost and storages}

\begin{table}[t]
\caption{Computational cost in FLOPs, parameters, and GPU memory usage for different optimizers in the model.
}
\label{tab:ab_computation}
\centering
\scalebox{0.75}{
\begin{tabular}{lcccccc}
\toprule
& \multicolumn{3}{c}{MAML}                     & \multicolumn{3}{c}{Meta-SGD} 
\\
\cmidrule(){2-4} \cmidrule(lr){5-7} 
Model                     & Extra FLOPs (M) & Extra parameters (M) &  Memory usage (G)  &  Extra FLOPs (M) & Extra parameters (M) &  Memory usage (G)  \\
\midrule
SGD 
&  0 &   0  &   7.3  &  0  &   0 &   8.4  
\\
Adam 
&  0.003  &   0.0004  &   7.7  &  0.004 &   0.0006 &   8.9  
\\
EMO
&  0.04 &   0.003  &   8.1  &  0.07 &   0.009 &   9.7
\\
\bottomrule
\end{tabular}}
\vspace{-1mm}
\end{table}

We  also report the extra computational cost and storage of different models with various optimizers  in Table~\ref{tab:ab_computation}. Although our model requires more parameters and computational costs compared to the baseline, it brings a 9.14\% improvement with MAML in accuracy on the \textit{mini}ImageNet. 

\paragraph{Analysis of episodic memory}
In this experiment, we meta-train 
MAML and MAML with EMO on the Bird dataset 
\begin{wrapfigure}{r}{0.4\linewidth}
\vspace{-1mm}
  \centering \includegraphics[width=0.7\linewidth] {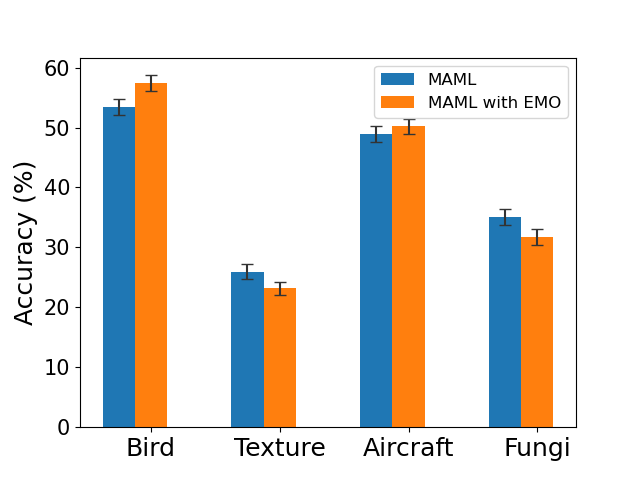}\\
    \caption{
    EMO is trained only on the Bird to show that EMO also holds semantic information. MAML with EMO  achieves better performance on the same dataset (Bird) and the similar shapes dataset (Aircraft), while it is harmful on the test tasks that have significant distribution shifts (Texture and Fungi) from the training tasks.} \label{fig:bird_acc}
  \vspace{-4mm}
\end{wrapfigure}
and meta-test on \textit{Meta-Dataset-BTAF}. Therefore the episodes saved in the memory 
are only from the Bird dataset.  The experiments that meta-train on the other three datasets are provided in Appendix~\ref{sec:emo_success}. From Figure~\ref{fig:bird_acc},  there is no doubt that MAML with EMO achieves a better performance than MAML on Bird. Surprisingly, MAML with EMO also outperforms MAML on Aircraft. It might be that the two datasets have more similar shapes (wings),  but Bird memory can still help accelerate the Aircraft tasks' training. However, when the test task has significant distribution shifts with training task, e.g., Texture, EMO harms performance. We will explore in future work how to use episodic memory to address cross-domain few-shot challenges.

\begin{table}[t]
\caption{Comparative results of different algorithms on \textit{Meta-Dataset-BTAF} and \textit{mini}ImageNet using  a Conv-4 backbone under the 5-way 1-shot setting. 
The results of the \textit{Meta-Dataset-BTAF}  of other methods are provided by~\citep{yao2019hierarchically, jiang2022subspace}.  Equipping ARML with EMO results in top-performance. }
\label{tab:plainmulti_res_1_shot}
\centering
\scalebox{0.9}{
\begin{tabular}{llccc|c}
\toprule
Method &  Bird & Texture & Aircraft & Fungi &\textit{mini}ImageNet \\\midrule
ANIL$^{\dagger}$~\citep{raghu2019rapid} & {53.36 \scriptsize{$\pm$ 1.42}} & {31.91 \scriptsize{$\pm$  1.25}} & {52.87 \scriptsize{$\pm$  1.34}} & {42.30 \scriptsize{$\pm$  1.28}} &  46.70 \scriptsize{$\pm$ 0.40}\\
MAML$^{\dagger}$~\citep{finn17} & 53.94 \scriptsize{$\pm$  1.45} & 31.66 \scriptsize{$\pm$ 1.31} & 51.37 \scriptsize{$\pm$  1.38} & 42.12 \scriptsize{$\pm$  1.36} &  48.70 \scriptsize{$\pm$ 1.84} \\
BMG~\citep{flennerhag2021bootstrapped} & {54.12 \scriptsize{$\pm$ 1.46}} & {32.19  \scriptsize{$\pm$  1.21}} & {52.09 \scriptsize{$\pm$  1.35}} & {43.00 \scriptsize{$\pm$  1.37}} & 52.97 \scriptsize{$\pm$ 0.85}\\
Meta-SGD$^{\dagger}$~\citep{li2017meta} & 55.58 \scriptsize{$\pm$ 1.43} & 32.38 \scriptsize{$\pm$ 1.32} & 52.99 \scriptsize{$\pm$ 1.36} & 41.74 \scriptsize{$\pm$  1.34} & 50.47 \scriptsize{$\pm$ 1.87}\\
MUSML~\citep{jiang2022subspace} & {60.52 \scriptsize{$\pm$ 0.33}} & \textbf{41.33} \scriptsize{$\pm$  1.30} & {54.69 \scriptsize{$\pm$  0.69}} & {{45.60} }\scriptsize{$\pm$  0.43} & -\\
HSML~\citep{yao2019hierarchically} & 60.98 \scriptsize{$\pm$ 1.50} & 35.01 \scriptsize{$\pm$  1.36} & {57.38} \scriptsize{$\pm$  1.40} & {44.02} \scriptsize{$\pm$  1.39} & 50.38  \scriptsize{$\pm$ 1.85}\\
TSA-MAML~\citep{zhou2021task} & {61.37 \scriptsize{$\pm$ 1.42}} & {35.41 \scriptsize{$\pm$  1.39}} & {{58.78} \scriptsize{$\pm$  1.37}} & {44.17} \scriptsize{$\pm$  1.25} &  48.44  \scriptsize{$\pm$ 0.91}\\ 
ARML~\citep{yao2020automated} & {{62.33} \scriptsize{$\pm$ 1.47}} & {35.65 \scriptsize{$\pm$  1.40}} & {{58.56} \scriptsize{$\pm$  1.41}} & {{44.82} \scriptsize{$\pm$  1.38}}  & 50.42  \scriptsize{$\pm$ 1.73}\\ 
\rowcolor{Gray}
\textbf{ARML with EMO} & \textbf{64.31} \scriptsize{$\pm$ 1.35} & {37.25 \scriptsize{$\pm$  1.43}} & \textbf{59.99} \scriptsize{$\pm$  1.35} & \textbf{46.15} \scriptsize{$\pm$  1.38} & \textbf{57.84} \scriptsize{$\pm$  0.93} \\
\bottomrule
\end{tabular}
}
\vspace{-1mm}
\end{table}

\paragraph{Comparison with the state-of-the-art}
We first compare our method on  \textit{Meta-Dataset-BTAF} using a Conv-4 backbone
under the 5-way 1-shot setting in Table~\ref{tab:plainmulti_res_1_shot}. In this comparison, we apply ARML~\citep{yao2020automated} with EMO to experiment since ARML is the current state-of-the-art algorithm based on optimization-based meta-learning. 
Our method achieves state-of-the-art performance on each dataset under the 5-way 1-shot setting. On Texture, our model surpasses the second best method, i.e., ARML~\citep{yao2020automated}, by $1.6\%$. The better performance confirms that EMO can find the most similar task to the test task and update the parameters so that it can converge faster and perform better. We also evaluate our method on traditional few-shot classiﬁcation, 
in which the training and test datasets are from the same dataset. {Note that the meta-training tasks of \textit{mini}imagenet have a slow, gradual shift in task distribution compared with meta-test task. In this experiment, we also apply ARML \citep{li2017meta} with EMO experiment.
The results have shown a significant improvement of 7.42\%. This suggests that the EMO technique is more effective even in the absence of task boundaries and a slow, gradual shift scenario.} The results demonstrate that optimization-based meta-learning benefits from EMO for traditional few-shot learning.

\section{Conclusions}
This paper introduces episodic memory optimization (EMO), which retains the gradient history of past experienced tasks in external memory. EMO accumulates long-term, general learning processes knowledge of past tasks, enabling it to learn new tasks quickly based on task similarity. Our experiments show that integrating EMO with several optimization-based meta-learning methods accelerates learning in all settings and datasets tested and improves their performance. We also prove that EMO with fixed-size memory converges under assumptions of strong convexity, regardless of which gradients are selected or how they are aggregated to form the update step. We conduct thorough ablation studies to demonstrate the effectiveness of the memory-augmented optimizer. Experiments on several few-shot learning datasets further substantiate the beneﬁt of the episodic memory optimizer.

\section*{Acknowledgment}
This work is financially supported by the Inception Institute of Artificial Intelligence, the University of Amsterdam and the allowance 
Top consortia for Knowledge and Innovation (TKIs) from the Netherlands Ministry of Economic Affairs and Climate Policy.

\bibliography{collas2023_conference}
\bibliographystyle{collas2023_conference}

\newpage
\appendix
\section{Appendix}

\section{Implementation Details}
\label{sec:app_imp}
We follow~\citep{finn17, yao2019hierarchically} by adopting the standard four-block convolutional layers as the feature extractor for our episodic memory optimizer and all baselines. 
We also conduct our experiments by ANIL~\citep{raghu2019rapid}, which  removes the inner-loop updates for the feature extractor network and applies inner-loop adaptation only to the classifier during training and testing.  
For all experiments, we keep the outer-loop optimizer consistent with the traditional optimization-based meta-learning approaches, e.g., Adam~\citep{Adam}. 
Our code will be publicly released.

\section{Hyperparameters \& Additional Experiment Settings}
{
The hyperparameters used in this paper are presented in Table~\ref{tab:maml_parameters}, ~\ref{tab:anil_parameters}, and ~\ref{tab:metasgd_parameters}. To implement MAML, we followed the approach of~\citep{finn17} by computing full Hessian-vector products. For the few-shot classification problem, we determine whether to increase clusters based on the change in averaged training accuracy.  Similar to~\citep{finn17}, we used a base learner with two hidden layers consisting of 40 neurons each, and the base learner comprised 4 standard convolutional blocks. All experiments were conducted using Tensorflow~\citep{abadi2016tensorflow}.}
\begin{table*}[ht]
\caption{Hyperparameter summary for MAML}
\label{tab:maml_parameters}
\begin{center}
\begin{tabular}{l|c|c}
\hline
Hyperparameters & miniImageNet & \textit{Meta-Dataset-BTAF} \\\hline
Input Scale (only for image data)  & $84\times84\times 3$ & $84\times84\times 3$ \\
Meta-batch Size (task batch size)  & 4 & 4\\
Inner loop learning rate ($\alpha$)  & 0.001 & 0.001\\
Outer loop learning rate ($\beta$) & 0 0.01 & 0.01\\
Filters of CNN (only for image data) & 32 & 32\\
Meta-training adaptation steps & 5 & 5\\
Task representation size  & 64 & 64\\
Image Embedding Size   & 64 & 64\\
Layer of transformer   & 6 & 6\\
Length of transformer for task representation (1-shot)   & 6  & 6 \\
Length of transformer for task representation (5-shot)   & 26 & 26\\
Number  selected memory (1-shot) & 20 & 20\\
Number  selected memory (5-shot) & 15 & 15\\
Length of transformer for $\texttt{Aggr}$  & 11 & 11\\
\hline
\end{tabular}
\end{center}
\end{table*}

\begin{table*}[ht]
\caption{Hyperparameter summary for ANIL}
\label{tab:anil_parameters}
\begin{center}
\begin{tabular}{l|c|c}
\hline
Hyperparameters & miniImageNet & \textit{Meta-Dataset-BTAF} \\\hline
Input Scale (only for image data)  & $84\times84\times 3$ & $84\times84\times 3$ \\
Meta-batch Size (task batch size)  & 8 & 8\\
Inner loop learning rate ($\alpha$)  & 0.001 & 0.001\\
Outer loop learning rate ($\beta$) & 0 0.01 & 0.01\\
Filters of CNN (only for image data) & 32 & 32\\
Meta-training adaptation steps & 5 & 5\\
Task representation size  & 64 & 64\\
Image Embedding Size   & 64 & 64\\
Layer of transformer   & 6 & 6\\
Length of transformer for task representation (1-shot)   & 6  & 6 \\
Length of transformer for task representation (5-shot)   & 26 & 26\\
Number  selected memory (1-shot) & 20 & 20\\
Number  selected memory (5-shot) & 15 & 15\\
Length of transformer for $\texttt{Aggr}$  & 11 & 11\\
\hline
\end{tabular}
\end{center}
\end{table*}

\begin{table*}[ht]
\caption{Hyperparameter summary for Meta-SGD}
\label{tab:metasgd_parameters}
\begin{center}
\begin{tabular}{l|c|c}
\hline
Hyperparameters & miniImageNet & \textit{Meta-Dataset-BTAF} \\\hline
Input Scale (only for image data)  & $84\times84\times 3$ & $84\times84\times 3$ \\
Meta-batch Size (task batch size)  & 4 & 4\\
Outer loop learning rate ($\beta$) & 0 0.01 & 0.01\\
Filters of CNN (only for image data) & 32 & 32\\
Meta-training adaptation steps & 3 & 3\\
Task representation size  & 64 & 64\\
Image Embedding Size   & 64 & 64\\
Layer of transformer   & 6 & 6\\
Length of transformer for task representation (1-shot)   & 6  & 6 \\
Length of transformer for task representation (5-shot)   & 26 & 26\\
Number  selected memory (1-shot) & 20 & 20\\
Number  selected memory (5-shot) & 15 & 15\\
Length of transformer for $\texttt{Aggr}$  & 11 & 11\\
\hline
\end{tabular}
\end{center}
\end{table*}

\section{Proof of Convergence}
\label{sec:proof}

To analyze the convergence rate of the model, we first derive the upper bound for the expectation $\mathbb{E} \norm{\Delta \theta_{t+1}}^2$ with respect to the independent random noises for all previous gradients $\{\epsilon_{j}\}_{j=1}^{t}$, where $\norm{ \cdot }$ is the spectral norm.
We reformulate the aggregation process of our method as a linear multi-step system. Thus the gradient for the $t$-th iteration is $\texttt{aggr}(\mathbf{g}_t, \mathcal{V}_t) {=} \sum_{s=0}^{S-1} w_{t, s} g_{t-s}$, where $S$ is the number of step in the system. 
By incorporating the aggregation process into the update rule Eq.~(\ref{eq:emo_inner}) and subtracting $\theta^*$ from both sides, we obtain the recursive formulation about the  difference $\Delta \theta_{t}$ as:
\begin{equation}
    \Delta \theta_{t+1} = \Delta \theta_{t} - \alpha \sum_{s=0}^{S-1} w_{t, s} g_{t-s}.
\end{equation}

In the paper, the gradient of each iteration is reformulated by adding its mean and the corresponding noise in Eq.~(\ref{eq:formulation of gradient}). For clarity in the proof below, we define the average rate of the gradient changes from the $t$-th iteration of model parameters to the optimal as:
\begin{equation}
\label{eq:average rate of the gradient changes}
    \mathcal{R}_t  = \frac{\nabla f(\theta_t) - \nabla f(\theta^*) }{\Delta \theta_t} = \int_{0}^{1} \nabla^2 f(\theta^* + u\Delta \theta_t)d u.
\end{equation}
With the assumptions about the objective function $f$, the average rate of gradient changes is also bounded between $\mu$ and $L$. By incorporating Eq.~(\ref{eq:average rate of the gradient changes}) into Eq.~(\ref{eq:formulation of gradient}), we simplify the recursive formulation about the difference $\Delta \theta_{t}$ as:
\begin{equation}
    \Delta \theta_{t+1} = \Delta \theta_{t} - \alpha \sum_{s=0}^{S-1} w_{t, s}  \mathcal{R}_{t-s} \Delta \theta_{t-s}  - \alpha \sum_{s=0}^{S-1} w_{t, s} \epsilon_{t-s}.
\end{equation}

We take recursive formulations about $\{\Delta \theta_{t+1-s}\}_{s=0}^{S-1}$ together and get the matrix version of the the recursion below:

\begin{equation}
\begin{aligned}
& ~~~~~~~~~~~~~~~~~~~~~~~~~~~~
\left[ \begin{array}
{c} 
\Delta \theta_{t+1} \\ 
\Delta \theta_{t} \\
\vdots \\
\Delta \theta_{t-S}
\end{array} \right] 
= A_t
\left[ \begin{array}
{c} 
\Delta \theta_{t} \\ 
\Delta \theta_{t-1} \\
\vdots \\
\Delta \theta_{t-S+1}
\end{array} \right] 
+
\left[ \begin{array}
{c} 
- \alpha \sum_{s=0}^{S-1} w_{t, s} \epsilon_{t-s} \\ 
0 \\
\vdots \\
0
\end{array} \right], \\
& \text{where}~A_t = \left[ \begin{array}
{ccccc} 
I -\alpha w_{t, 0} \mathcal{R}_{t} & -\alpha w_{t, 1} \mathcal{R}_{t-1} & \cdots & -\alpha w_{t, S-2} \mathcal{R}_{t-S+2} & -\alpha w_{t, S-1} \mathcal{R}_{t-S+1}\\ 
I & 0 & \cdots  & 0 & 0 \\
0 & I & \cdots  & 0 & 0 \\
\vdots & \vdots & \ddots & \vdots & \vdots \\
0 & 0 & \cdots  & I & 0 \\
\end{array} \right]. 
\end{aligned}
\label{eq:matrix vertion}
\end{equation}

Note that $A_t$ is the system matrix at the $t$-th iteration. By unrolling the recursion below, the upper bound of the expectation $\mathbb{E} \norm{\Delta \theta_{t+1}}^2$ can be derived :
\begin{equation}
\begin{aligned}
\mathbb{E} \norm{\Delta \theta_{t+1}}^2 
& \leq \mathbb{E}_{\epsilon_t, \cdots, \epsilon_1} 
\norm{\left[ \begin{array}
{c} 
\Delta \theta_{t+1} \\ 
\Delta \theta_{t} \\
\vdots \\
\Delta \theta_{t-S}
\end{array} \right] }^2 \\
& = \mathbb{E}_{\epsilon_t, \cdots, \epsilon_1} \norm{A_t
\left[ \begin{array}
{c} 
\Delta \theta_{t} \\ 
\Delta \theta_{t-1} \\
\vdots \\
\Delta \theta_{t-S+1}
\end{array} \right] 
+
\left[ \begin{array}
{c} 
- \alpha \sum_{s=0}^{S-1} w_{t, s} \epsilon_{t-s} \\ 
0 \\
\vdots \\
0
\end{array} \right] }^2 \\
& = \norm{A_t}^2 
\mathbb{E}_{\epsilon_{t-1}, \cdots, \epsilon_1} 
\norm{\left[ \begin{array}
{c} 
\Delta \theta_{t} \\ 
\Delta \theta_{t-1} \\
\vdots \\
\Delta \theta_{t-S+1}
\end{array} \right] }^2 + \alpha^2 \sum_{s=0}^{S} w_{t, s}^2  \mathbb{E}_{\epsilon_{t-s}} \norm{ \epsilon_{t-s}}^2 \\
& \leq \norm{A_t}^2 
\mathbb{E}_{\epsilon_{t-1}, \cdots, \epsilon_1} 
\norm{\left[ \begin{array}
{c} 
\Delta \theta_{t} \\ 
\Delta \theta_{t-1} \\
\vdots \\
\Delta \theta_{t-S+1}
\end{array} \right] }^2 + \alpha^2 S \sigma^2 \\
& \cdots \\
& \leq \prod_{j=1}^{t} \norm{A_j}^2 \norm{\Delta \theta_1}^2 + \alpha^2 S \sigma^2 \sum_{j=1}^{S} (\norm{A_t}^2 \cdots \norm{A_{j+1}}^2).
\end{aligned}
\label{eq:upper bound}
\end{equation}

According to the definition of the spectral norm and the properties of block matrix~\citep{polyak1964some, assran2020convergence, mcrae2021memory}, we get the upper bound of the spectral norm below:
\begin{equation}
\begin{aligned}
&~~~~~~~~~~~~~~~~~~~~~~~~~~~~~~~~~~~~~~~~~~~~~~~~~~~~~~~~~~~~~~
\norm{A_t} \leq \lambda_{t}(\widehat{A}_t^{\top}\widehat{A}_t), \\ & \text{where}~\widehat{A}_t = 
\left[ \begin{array} {ccccc} 
1 -\alpha w_{t, 0}{\tau}_{t} & -\alpha w_{t, 1} {\tau}_{t-1} & \cdots & -\alpha w_{t, S-2} {\tau}_{t-S+2} & -\alpha w_{t, S-1} {\tau}_{t-S+1}\\ 
1 & 0 & \cdots  & 0 & 0 \\
0 & 1 & \cdots  & 0 & 0 \\
\vdots & \vdots & \ddots & \vdots & \vdots \\
0 & 0 & \cdots  & 1 & 0 \\
\end{array} \right].
\end{aligned}
\label{eq:spectral radius}
\end{equation}

Note that $\lambda_{t}(\widehat{A}_t^{\top}\widehat{A}_t)$ is the square root of the largest eigenvalue of the matrix $\widehat{A}_t^{\top}\widehat{A}_t$. The matrix $\widehat{A}_t \in \mathbb{R}^{S\times S}$ has bounded hyperparameters:  ${\tau}_t \in [\mu, L]$ and $ w_t \in [0, 1]$. We introduce $\lambda_{\max}$ as the upper bound for all $\lambda_{t}$ corresponding to all system matrices.
Since the learning rate is chosen sufficiently small such that $\lambda_{\max} < 1$, we further simplify Eq.~(\ref{eq:upper bound}) below:
\begin{equation}
    \mathbb{E} \norm{\Delta \theta_{t+1}}^2  \leq \lambda_{\max}^{2t}\norm{\Delta \theta_1}^2 + \frac{\alpha^2 \sigma^2 S}{1-\lambda^2_{\max}}.  
\end{equation}

Recall that $f(\cdot)$ is assumed to be $L$-smooth, 
we get the convergence rate of our model as
\begin{equation}
    f(\theta_{t+1}) - f(\theta^*) \leq \frac{L}{2} (\lambda_{\max}^{2t}\norm{\Delta \theta_1}^2 + \frac{\alpha^2  \sigma^2 S}{1-\lambda_{\max}^2}).
\end{equation}

\section{Algorithm}
{
This section introduces the MAML with EMO for Few-shot learning, denoted as MAML with EMO. The algorithm for the meta-training and meta-test, is presented in Algorithms~\ref{alg:EMO_train} and~\ref{alg:EMO_test}. 
} 
\begin{algorithm}[ht]
\small
\caption{MAML with EMO for few-shot meta-learning (meta-train)}
\label{alg:EMO_train}
\begin{algorithmic}
\REQUIRE $p(\mathcal{T})$: distribution over tasks
\REQUIRE $\alpha$, $\beta$: step size hyperparameters
\STATE randomly initialize $\theta$
\WHILE{not done}
\STATE Sample batch of tasks $\mathcal{T}_t \sim p(\mathcal{T})$
  \FORALL{$\mathcal{T}_t$}
      \STATE Sample $K$ datapoints $\mathcal{S}_t=\{\mathbf{x}^{(j)}, y^{(j)}\}$ from $\mathcal{T}_i$
      \STATE Evaluate $\mathbf{g}_t = \nabla_\theta \mathcal{L}(f_\theta)$ using $\mathcal{D}$ and $\mathcal{L}$
     \STATE Compute the task representation $K_t$ by Eq.~\ref{eqn:key}
      \STATE Retrieve gradient $\mathcal{V}_t$ from memory based on the task similarity. 
      \STATE Compute adapted parameters with gradient descent: $\theta'=\theta-\alpha \cdot \texttt{Aggr}(\mathbf{g}_t, \mathcal{V}_t)$
      \STATE Update memory content $M_c = \texttt{Controller}(\mathbf{g}_t, \hat{M}_c)$, where $\hat{M}_c$ is selected memory to be replaced
      \STATE Sample data points $\mathcal{Q}_t=\{\mathbf{x}^{(j)}, y^{(j)}\}$ from $\mathcal{T}_t$ for the meta-update
 \ENDFOR
 \STATE Update $\theta \leftarrow \theta - \beta \nabla_\theta \sum_{\mathcal{T}_i \sim p(\mathcal{T})}  \mathcal{L} ( f_{\theta_i'})$ using each $\mathcal{Q}_t'$ and $\mathcal{L}$ 
\ENDWHILE
\end{algorithmic}
\end{algorithm}

\begin{algorithm}[ht]
\small
\caption{MAML with EMO for few-shot meta-learning (meta-test)}
\label{alg:EMO_test}
\begin{algorithmic}
\REQUIRE $\mathcal{T}^{ts}$: meta-test task
\REQUIRE $\alpha$: inner step size hyperparameter, $\theta$: meta-learned parameter

\STATE Sample $K$ datapoints $\mathcal{S}_t=\{\mathbf{x}^{(j)}, y^{(j)}\}$ from $\mathcal{T}^{ts}$
\STATE Evaluate $\mathbf{g}_t = \nabla_\theta \mathcal{L}(f_\theta)$ using $\mathcal{D}$ and $\mathcal{L}$
\STATE Compute the task representation $K_t$ by Eq.~\ref{eqn:key}
\STATE Retrieve gradient $\mathcal{V}_t$ from memory based on the task similarity. 
\STATE Compute adapted parameters with gradient descent: $\theta'=\theta-\alpha \cdot \texttt{Aggr}(\mathbf{g}_t, \mathcal{V}_t)$
\STATE Sample datapoints $\mathcal{Q}^{ts}=\{\mathbf{x}^{(j)}, y^{(j)}\}$ from $\mathcal{T}^{ts}$ for evaluation
\RETURN $\hat{y}^j = f_{\theta^{'}}(\mathbf{x}^j)$
\end{algorithmic}
\end{algorithm}

\section{More results}
\subsection{Comparison with other optimizers}
\label{sec:other_optimizers}
\begin{table}[ht]
\caption{Comparison with other optimizers on \textit{Meta-Dataset-BTAF} under the 5-way 1-shot setting. EMO achieves better performance compared to other optimizers on all datasets.} 
\label{tab:optimizers}
\centering
\scalebox{0.9}{
\begin{tabular}{lcccc}
\toprule
& \multicolumn{4}{c}{MAML} \\
\cmidrule(lr){2-5} 
Dataset & w/ SGD & w/ Momentum & w/ Adam & w/ EMO\\
\midrule
Bird & \textbf{53.94} \scriptsize{$\pm$  1.45}&52.98 \scriptsize{$\pm$   1.42} & 52.55 \scriptsize{$\pm$  1.41} & \textbf{56.32} \scriptsize{$\pm$  1.33} \\
Texture &31.66 \scriptsize{$\pm$  1.31} &31.38 \scriptsize{$\pm$  1.31} & 30.95 \scriptsize{$\pm$  1.34}&\textbf{34.75} \scriptsize{$\pm$   1.41} \\
Aircraft &\textbf{51.37} \scriptsize{$\pm$   1.38} &51.09 \scriptsize{$\pm$   1.35} &50.15 \scriptsize{$\pm$  1.33} &  \textbf{53.99} \scriptsize{$\pm$   1.33}\\
Fungi &\textbf{42.12} \scriptsize{$\pm$   1.36} & \textbf{41.54} \scriptsize{$\pm$   1.35} & \textbf{41.04} \scriptsize{$\pm$   1.31} &\textbf{43.15} \scriptsize{$\pm$   1.36} \\
\bottomrule
\end{tabular}}
\end{table}
To show the benefit of the episodic memory optimizer, we compare MAML~\citep{finn17}, Meta-SGD~\citep{li2017meta}, and ANIL~\citep{raghu2019rapid} with their EMO variants, where each variant uses EMO as the inner-loop optimizer. Table~\ref{tab:optimizers} shows each method with EMO achieves better performance by a large margin than the original methods on four different datasets. More importantly, the most challenging, which has the largest domain gap Texture, delivers 34.75\%, surpassing the Meta-SGD by 2.09\%. We attribute the improvements to our model’s ability to leverage the episodic memory to adjust the model parameters, allowing the model to update the test task model using the most training task-like update, and thus leading to improvements over original models.

\subsection{Effect of different aggregation functions}
\label{sec:ANIL_agg}
We also give the ANIL with EMO for ablating the effect of EMO’s aggregation function used to compute the new gradients. We report the performance of ANIL  with EMO using different aggregation functions in Table~\ref{tab:ANIL_agg}. The best-suited aggregation function for ANIL with EMO is the \texttt{Transformer}. To ensure consistency of implementation on each dataset, we choose the \texttt{Transformer} aggregation function for ANIL with EMO.
\begin{table*}[ht]

\caption{Effect of ANIL with different aggregation functions. \texttt{Mean} achieves better performance than alternatives. } 
\label{tab:ANIL_agg}
\centering
\scalebox{0.9}{\begin{tabular}{lccc}
\toprule
& \multicolumn{3}{c}{ANIL with EMO} \\
\cmidrule(lr){2-4} 
Dataset & \texttt{sum} & \texttt{Mean} & \texttt{Transformer} \\
\midrule
Bird & 54.91 \scriptsize{$\pm$   1.33} & {55.18} \scriptsize{$\pm$   1.34} & {54.78} \scriptsize{$\pm$  1.33} 
\\
Texture & 32.71 \scriptsize{$\pm$   1.30} & {33.14} \scriptsize{$\pm$   1.40} & {33.15} \scriptsize{$\pm$   1.41}\\
Aircraft & {53.16} \scriptsize{$\pm$   1.40} & 52.11 \scriptsize{$\pm$   1.38} & {52.79} \scriptsize{$\pm$   1.33}\\
Fungi & 43.17 \scriptsize{$\pm$   1.34} & {43.07} \scriptsize{$\pm$   1.31} &  {43.75} \scriptsize{$\pm$   1.36}\\
\bottomrule
\end{tabular}}
\end{table*}

\subsection{Effect of memory controller}
\label{sec:ANIL_controller}
We further assess the effect of the memory controller with ANIL with EMO and Meta-SGD with EMO in Table~\ref{tab:ANIL_controller}. With \texttt{CLOCK-EM}, Meta-SGD with EMO achieves better performance on all datasets, while  ANIL with EMO leads to a small but consistent gain under all the datasets with \texttt{LRU-EM}. To ensure consistency of implementation on each dataset, we choose the \texttt{LRU-EM} function for ANIL with EMO, and \texttt{CLOCK-EM} is used for Meta-SGD with EMO.
\begin{table*}[ht]

\caption{Effect of ANIL with different memory controllers. \texttt{LRU-EM} achieves better performance than alternatives. } 
\label{tab:ANIL_controller}
\centering
\scalebox{0.9}{\begin{tabular}{lccc}
\toprule
& \multicolumn{3}{c}{ANIL with EMO} \\
\cmidrule(lr){2-4} 
Dataset & \texttt{FIFO-EM} & \texttt{CLOCK-EM} & \texttt{LRU-EM} \\
\midrule
Bird & 50.11 \scriptsize{$\pm$   1.31} &{53.91} \scriptsize{$\pm$   1.34} & \textbf{54.78} \scriptsize{$\pm$  1.43} 
\\
Texture & 29.11 \scriptsize{$\pm$   1.41} & {32.94} \scriptsize{$\pm$   1.40} & \textbf{33.15} \scriptsize{$\pm$   1.31}\\
Aircraft & 47.96 \scriptsize{$\pm$   1.40} & \textbf{53.91} \scriptsize{$\pm$   1.35} & {52.79} \scriptsize{$\pm$   1.33}\\
Fungi & 40.97 \scriptsize{$\pm$   1.35} & {43.17} \scriptsize{$\pm$   1.35} &  \textbf{43.75} \scriptsize{$\pm$   1.31}\\
\bottomrule
\end{tabular}}
\end{table*}

\begin{table*}[t]

\caption{Effect of Meta-SGD with different memory controllers. \texttt{LRU-EM} achieves better performance than alternatives. } 
\label{tab:MSGD_controller}
\centering
\scalebox{0.9}{\begin{tabular}{lccc}
\toprule
& \multicolumn{3}{c}{Meta-SGD with EMO} \\
\cmidrule(lr){2-4} 
Dataset & \texttt{FIFO-EM} & \texttt{CLOCK-EM} & \texttt{LRU-EM} \\
\midrule
Bird & 53.05 \scriptsize{$\pm$   1.34} & \textbf{58.95} \scriptsize{$\pm$   1.41} & {57.31} \scriptsize{$\pm$  1.34} 
\\
Texture & 32.13 \scriptsize{$\pm$   1.41} & \textbf{36.26} \scriptsize{$\pm$   1.33} & {35.95} \scriptsize{$\pm$   1.41}\\
Aircraft & 49.16 \scriptsize{$\pm$   1.41} & 55.21 \scriptsize{$\pm$   1.35} & \textbf{56.19} \scriptsize{$\pm$   1.34}\\
Fungi & 41.61 \scriptsize{$\pm$   1.34} & \textbf{45.24} \scriptsize{$\pm$   1.35} &  {44.75} \scriptsize{$\pm$   1.36}\\
\bottomrule
\end{tabular}}
\end{table*}

\subsection{Analysis of episodic memory}
\label{sec:emo_success}
\begin{figure} [ht]
  \begin{minipage}{1.0\columnwidth}
         \subfloat[Trained only Texture]{
      \includegraphics[width=.33\columnwidth]{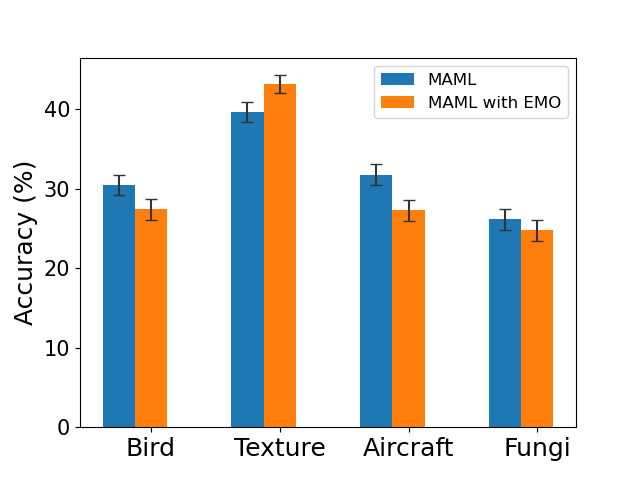}}
       \subfloat[Trained only Aircraft]{
      \includegraphics[width=.33\columnwidth]{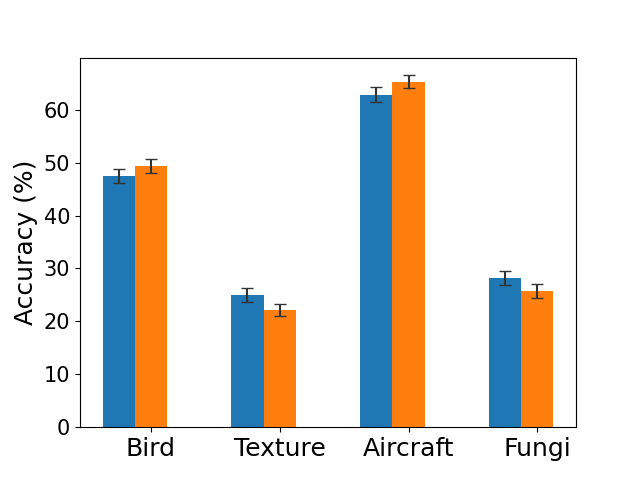}}
       \subfloat[Trained only Fungi]{
      \includegraphics[width=.33\columnwidth]{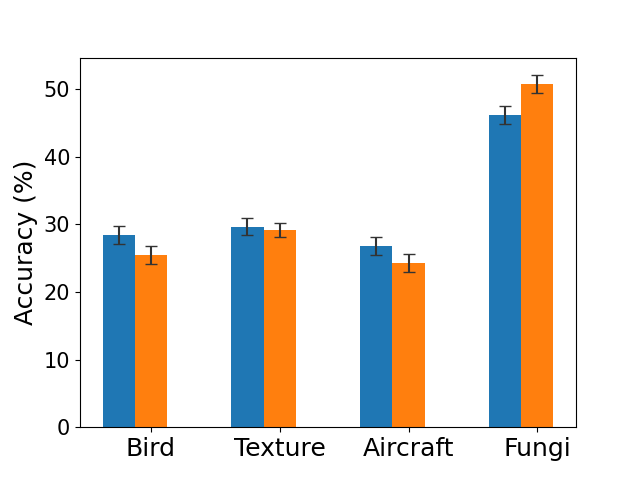}}
\end{minipage}
\caption{Analysis of episodic memory.}
	\label{fig:emo_success}

\end{figure}

In this section, we further analysis of our proposed episodic memory with the other three datasets.  
In this experiment, we meta-train MAML and MAML with EMO on the Texture, Aircraft, and Fungi datasets, respectively, and meta-test on \textit{Meta-Dataset-BTAF}. Therefore the episodes saved in the memory are from the  Texture, Aircraft, and Fungi, respectively. The results are shown in Figure~\ref{fig:emo_success}. Consistent with the results in the Figure~\ref{fig:bird_acc}, MAML with EMO has a significant performance improvement when the meta-training dataset is the same as the meta-test dataset. Interestingly, the memory of Aircraft can also help Bird to achieve better performance in Figure~\ref{fig:emo_success} (b). Similarly, when the test task has large distribution shifts with the training task, the memory will not be useful or even harmful. 

\subsection{Adaptation speed}
\label{sec:speed}
\begin{figure} [ht] 
\centering
      \includegraphics[width=.6\columnwidth]{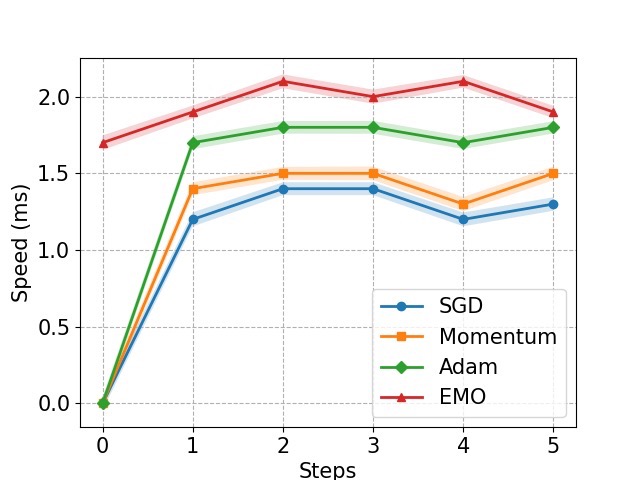}
\caption{Adaptation speed of MAML with different optimizers at each inner-loop step.}
	\label{fig:speed}
\end{figure}
In addition to our findings on model performance, we have also examined the adaptation speed of different optimizers at each inner-loop step. We found that while our model requires a higher adaptation speed of 1.7 for EMO due to the need for initialization gradients from memory at step 0, the other optimizers have a speed of 0 since they do not need to compute gradients. Despite this difference in adaptation speed, our model was able to converge quickly and consistently achieve the highest accuracy at each step (see Figure~\ref{fig:steps}). Overall, these results suggest that our model is a promising approach for achieving both high accuracy and efficient adaptation speed in few-shot learning tasks.

\subsection{Effect of the number of keys}
\label{sec:key}
\begin{figure} [t] 
\centering
      \includegraphics[width=.6\columnwidth]{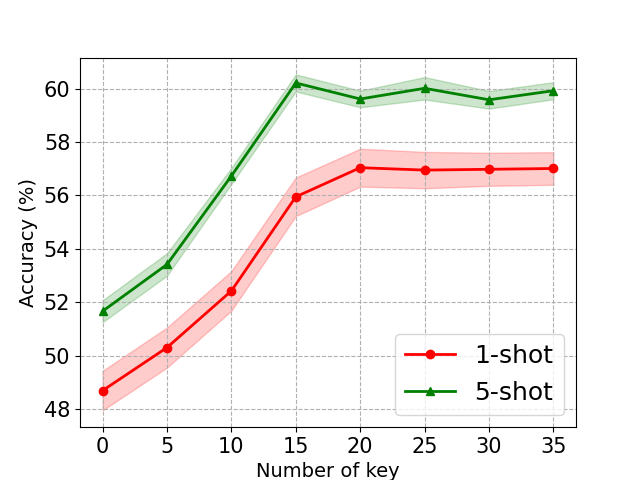}
\caption{Effect of the number of keys for the MAML with EMO.}
	\label{fig:key}
\end{figure}
{
The choice of the number of keys for the selected memory is an important hyperparameter to consider when implementing the MAML with the EMO approach for few-shot classification. From figure~\ref{fig:key}, we observed that as k increases, the performance of MAML with EMO also increases for 1-shot tasks, and this trend continues until k reaches 20, at which point the performance converges. For 5-shot tasks, we found that a k value of 15 achieves the best performance. We also set k=20 and k=15 in the Sota experiments, respectively.
These results highlight the importance of carefully tuning the hyperparameters to achieve optimal performance in few-shot classification tasks, and suggest that the choice of k may be task-specific.}

\begin{table*}[t]

\caption{{Comparison with different gradient aggregation functions.}} 
\label{tab:gradient_agg}
\centering
\scalebox{0.9}{\begin{tabular}{lcc}
\toprule
& \multicolumn{2}{c}{MAML with EMO} \\
\cmidrule(lr){2-3} 
Dataset &  Mean-based gradient & Transformer-based gradient \\
\midrule
Bird & 56.32 \scriptsize{$\pm$   1.33} & {57.05} \scriptsize{$\pm$   1.30}
\\
Texture & 34.75 \scriptsize{$\pm$   1.41} & {35.23} \scriptsize{$\pm$   1.39} \\
Aircraft & {53.99} \scriptsize{$\pm$   1.33} & 54.73 \scriptsize{$\pm$   1.35} \\
Fungi & 43.15 \scriptsize{$\pm$   1.36} & {43.45} \scriptsize{$\pm$   1.37} \\
\bottomrule
\end{tabular}}
\end{table*}

\subsection{Comparison with the state-of-the-art on few-shot learning datasets}
\label{sec:sota}
We further conduct experiments on the \textit{Meta-Dataset-BTAF} and \textit{mini}ImageNet under  the 5-way 5-shot setting in Table~\ref{tab:plainmulti_res_5_shot}. We also give the  comparative results for few-shot learning on \textit{mini}ImageNet and \textit{tired}ImageNet using a ResNet-12 back in the Table~\ref{tab:mini_resnet}. In these comparison, we apply ARML~\citep{yao2020automated} with EMO  to do the experiment. Our method achieves state-of-the-art performance on all  benchmarks under the 5-way 5-shot setting.  
\label{sec:app_5-shot}
\begin{table*}[t]
\caption{Comparative results of different algorithms on the \textit{Meta-Dataset-BTAF}  using  a Conv-4 backbone under the 5-way 5-shot setting. The results of other methods are provided by~\citep{yao2019hierarchically, jiang2022subspace}. Equipping ARML with EMO makes it a consistent top-performer.}
\label{tab:plainmulti_res_5_shot}
\centering
\scalebox{0.85}{
\begin{tabular}{lcccc|c}
\toprule
 Method &  Bird & Texture & Aircraft & Fungi & \textit{mini}ImageNet\\\midrule
 MAML~\citep{finn17} & 68.52 \scriptsize{$\pm$ 0.79} & 44.56 \scriptsize{$\pm$ 0.68} & 66.18 \scriptsize{$\pm$  0.71} & 51.85 \scriptsize{$\pm$ 0.85} & 63.11 \scriptsize{$\pm$ 0.92} \\
Meta-SGD~\citep{li2017meta} & 67.87 \scriptsize{$\pm$ 0.74} & 45.49 \scriptsize{$\pm$ 0.68} & 66.84 \scriptsize{$\pm$ 0.70} & 52.51 \scriptsize{$\pm$ 0.81} & 64.03 \scriptsize{$\pm$ 0.94} \\
HSML~\citep{yao2019hierarchically} & 71.68 \scriptsize{$\pm$  0.73} & 48.08 \scriptsize{$\pm$  0.69} & 73.49 \scriptsize{$\pm$  0.68} & 56.32 \scriptsize{$\pm$  0.80} & 65.91 \scriptsize{$\pm$ 0.95} \\
ARML~\citep{yao2020automated} & 73.34 \scriptsize{$\pm$ 0.70} & {49.67 \scriptsize{$\pm$  0.67}} & {74.88 \scriptsize{$\pm$  0.64}} & {57.55 \scriptsize{$\pm$  0.82}} & 66.87 \scriptsize{$\pm$ 0.93} \\
TSA-MAML~\citep{zhou2021task} & {72.31 \scriptsize{$\pm$ 0.71}} & {49.50 \scriptsize{$\pm$  0.68}} & {74.01 \scriptsize{$\pm$  0.70}} & {56.95 \scriptsize{$\pm$  0.80}} & 65.52 \scriptsize{$\pm$ 0.92} \\
ANIL~\citep{raghu2019rapid} & {70.67\scriptsize{$\pm$ 0.72}} & {44.67 \scriptsize{$\pm$  0.95}} & {66.05 \scriptsize{$\pm$  1.07}} & {52.89 \scriptsize{$\pm$  0.30}} & 61.50 \scriptsize{$\pm$ 0.92} \\
BMG~\citep{flennerhag2021bootstrapped} & {71.56 \scriptsize{$\pm$ 0.76}} & {49.44  \scriptsize{$\pm$  0.73}} & {66.83 \scriptsize{$\pm$  0.79}} & {52.56 \scriptsize{$\pm$  0.89}} & 66.73 \scriptsize{$\pm$ 0.91} \\
MUSML~\citep{jiang2022subspace} & {76.69 \scriptsize{$\pm$ 0.72}} & 52.41 \scriptsize{$\pm$  0.75} & {77.76 \scriptsize{$\pm$  0.82}} & {57.74 \scriptsize{$\pm$  0.81}} & 65.12 \scriptsize{$\pm$ 1.48} \\
\rowcolor{Gray}
\textbf{ARML with EMO} & \textbf{77.17} \scriptsize{$\pm$ 0.65} & \textbf{53.25} \scriptsize{$\pm$  0.68} & \textbf{77.83} \scriptsize{$\pm$  0.63} & \textbf{59.15} \scriptsize{$\pm$ 0.79} & \textbf{71.05} \scriptsize{$\pm$ 0.91}\\

\bottomrule
\end{tabular}}
\end{table*}

\begin{table*}[t]
\caption{Comparative results for few-shot learning on  \textit{mini}Imagenet and \textit{tiered}Imagenet using a ResNet-12 backbone.  ARML with EMO can also improve performance for traditional few-shot learning.}

\centering
\scalebox{0.90}{
\begin{tabular}{lcccc}
\toprule
 &  \multicolumn{2}{c}{\textbf{\textit{mini}Imagenet 5-way}} & \multicolumn{2}{c}{\textbf{\textit{tiered}Imagenet 5-way}}  \\
\textbf{Method} & \textbf{1-shot} & \textbf{5-shot} & \textbf{1-shot} & \textbf{5-shot}  \\
\midrule

SNAIL \citep{mishra2018simple}  & 55.71 \scriptsize{$\pm$ 0.99} & 68.88 \scriptsize{$\pm$ 0.92} & - & - \\
Dynamic FS \citep{gidaris18}  & 55.45 \scriptsize{$\pm$ 0.89} & 70.13 \scriptsize{$\pm$ 0.68} & - & - \\
TADAM \citep{oreshkin2018tadam}  & 58.50 \scriptsize{$\pm$ 0.30} & 76.70 \scriptsize{$\pm$ 0.30} & - & - \\
MTL \citep{sun19}  & 61.20 \scriptsize{$\pm$ 1.80} & 75.50 \scriptsize{$\pm$ 0.80} & - & - \\
VariationalFSL \citep{zhang19}  & 61.23 \scriptsize{$\pm$ 0.26} & 77.69 \scriptsize{$\pm$ 0.17} & - & - \\
TapNet \citep{yoon19}  & 61.65 \scriptsize{$\pm$ 0.15} & 76.36 \scriptsize{$\pm$ 0.10} & 63.08 \scriptsize{$\pm$ 0.15} & 80.26 \scriptsize{$\pm$ 0.12} \\
MetaOptNet \citep{lee19}  & 62.64 \scriptsize{$\pm$ 0.61} & 78.63 \scriptsize{$\pm$ 0.46} & 65.81 \scriptsize{$\pm$ 0.74} & 81.75 \scriptsize{$\pm$ 0.53} \\
CTM \citep{li19}  & 62.05 \scriptsize{$\pm$ 0.55} & 78.63 \scriptsize{$\pm$ 0.06} & 64.78 \scriptsize{$\pm$ 0.11} & 81.05 \scriptsize{$\pm$ 0.52} \\
CAN \citep{hou19}  & 63.85 \scriptsize{$\pm$ 0.48} & 79.44 \scriptsize{$\pm$ 0.34} & {69.89} \scriptsize{$\pm$ 0.51} & {84.23} \scriptsize{$\pm$ 0.37} \\

HVM \citep{du2021hierarchical}   & {67.83} \scriptsize{$\pm$ 0.57} & {83.88} \scriptsize{$\pm$ 0.51} &  {73.67} \scriptsize{$\pm$ 0.71} &  {88.05} \scriptsize{$\pm$ 0.44} \\
\rowcolor{Gray}
{\textbf{ARML with EMO}}  & $\textbf{69.15}$ \scriptsize{$\pm$ 0.34} & $\textbf{84.13}$ \scriptsize{$\pm$ 0.25} & $\textbf{75.17}$ \scriptsize{$\pm$ 0.35} &  $\textbf{89.05}$ \scriptsize{$\pm$ 0.25}
\\

\bottomrule
\end{tabular}}
\label{tab:mini_resnet}

\end{table*}

\end{document}